\documentclass[sigconf]{aamas} 

\usepackage{balance} 

\usepackage{algorithm}
\usepackage{algpseudocode}
\usepackage{graphicx}
\usepackage{amsthm}
\usepackage{textcomp}
\usepackage{xcolor}
\usepackage{gensymb}

\usepackage{diagbox}
\usepackage{caption}
\usepackage{subcaption}
\usepackage{tabularx}
\usepackage{comment}
\usepackage{tikz}
\usepackage{pgfplots}
\pgfplotsset{compat=1.18}

\usepackage{enumitem}

\setcopyright{none}
\acmConference[AAMAS '26]{This is an extended version of the paper published in the Proceedings of the 25th International Conference on Autonomous Agents and Multiagent Systems (AAMAS 2026). The official version of record is available at \url{https://doi.org/10.65109/CZPZ7833}.}
\copyrightyear{}
\acmYear{}
\acmDOI{}
\acmPrice{}
\acmISBN{}

\acmSubmissionID{23}

\title[Fairness over Equality]{Fairness over Equality: Correcting Social Incentives in Asymmetric Sequential Social Dilemmas}

\author{Alper Demir}
\affiliation{
  \institution{Izmir University of Economics \\ \& University of Edinburgh}
  \city{Izmir}
  \country{Turkey}}
\email{alper.demir@ieu.edu.tr}

\author{Hüseyin Aydın}
\affiliation{
  \institution{Middle East Technical University \\ \& Utrecht University}
  \city{Ankara}
  \country{Turkey}}
\email{huseyin@ceng.metu.edu.tr}

\author{Kale-ab Abebe Tessera}
\affiliation{
  \institution{University of Edinburgh}
  \city{Edinburgh}
  \country{United Kingdom}}
\email{k.tessera@ed.ac.uk}

\author{David Abel}
\affiliation{
  \institution{University of Edinburgh}
  \city{Edinburgh}
  \country{United Kingdom}}
\email{david.abel@ed.ac.uk}

\author{Stefano V. Albrecht}
\affiliation{
  \institution{DeepFlow}
  \city{London}
  \country{United Kingdom}}
\email{stefano.albrecht@deepflow.com}

\begin{abstract}
Sequential Social Dilemmas (SSDs) provide a key framework for studying how cooperation emerges when individual incentives conflict with collective welfare. In Multi-Agent Reinforcement Learning, these problems are often addressed by incorporating intrinsic drives that encourage prosocial or fair behavior. However, most existing methods assume that agents face identical incentives in the dilemma and require continuous access to global information about other agents to assess fairness. In this work, we introduce asymmetric variants of well-known SSD environments and examine how natural differences between agents influence cooperation dynamics. Our findings reveal that existing fairness-based methods struggle to adapt under asymmetric conditions by enforcing raw equality that wrongfully incentivize defection. To address this, we propose three modifications: (i) redefining fairness by accounting for agents’ reward ranges, (ii) introducing an agent-based weighting mechanism to better handle inherent asymmetries, and (iii) localizing social feedback to make the methods effective under partial observability without requiring global information sharing. Experimental results show that in asymmetric scenarios, our method fosters faster emergence of cooperative policies compared to existing approaches, without sacrificing scalability or practicality.
\end{abstract}


\newcommand{\BibTeX}{\rm B\kern-.05em{\sc i\kern-.025em b}\kern-.08em\TeX}

\newtheorem{definition}{Definition}
\newcommand{\figref}{Fig.~}
 
\algnewcommand{\LineComment}[1]{\State \(\triangleright\) #1}

\begin{document}


\pagestyle{fancy}
\fancyhead{}


\maketitle 


\section{Introduction}

Multi-Agent Reinforcement Learning (MARL) addresses problems involving multiple agents, each with potentially distinct preferences and policies \cite{marl-book}. The dynamics of these interactions require agents to navigate not only the environment but also the actions of their counterparts, rendering MARL problems highly non-stationary \cite{papoudakis2019dealing}. This complexity is notably amplified when navigating situations where individual and group objectives diverge. Social dilemmas such as Prisoner’s Dilemma are examples of such games where agents’ individual benefit are in tension with the groups’ overall reward. In MARL, these types of problems are temporally extended, leading to Sequential Social Dilemmas (SSDs) \cite{leibo2017multi}. The literature has predominantly proposed solutions centered around cultivating intrinsic motivations that drive agents toward prosocial behavior, thereby promoting actions conducive to group welfare \cite{eccles2019learning, hughes2018inequity, PeysakhovichL18}. Alternatively, studies explore agents incentivizing one another toward prosocial behavior \cite{hostallero2020inducing,lupu2020gifting,yang2020learning}. 

Fairness-based intrinsic motivation methods \cite{hughes2018inequity,McKeeGMDHL20} have shown promise in encouraging cooperation in multi-agent systems. These approaches typically assume that agents are identical, sharing the same capabilities, roles, and reward structures, and thus interpret differences in rewards as differences in cooperative behavior. While this assumption holds in symmetric environments such as the classic Prisoner's Dilemma, it breaks down in more realistic, asymmetric settings in which agents are not identical. Under such asymmetries, even identifying Pareto-optimal joint outcomes becomes non-trivial, since the feasible payoff surface itself differs across agents with unequal reward ranges \cite{christianos2023pareto}.

In \textit{asymmetric} SSDs, agents may differ in physical capabilities, access to environmental resources, or reward structures, leading to unequal opportunities and outcomes. This disparity can come about even when agents behave identically, highlighting the need for fairness methods that handle such asymmetries. For example, consider the design of a tax system. Requiring every individual to pay the same flat amount may appear equal but could be seen as inherently unfair, as it places a disproportionately heavy burden on those with lower incomes. A more fair approach accounts for individuals’ capacity to pay—such as through progressive taxation, where those who earn more contribute more. Beyond income, individuals with greater influence on the broader society are often expected to contribute more to the common good, whether through higher taxes, leadership responsibilities, or adherence to social norms. Inspired by these principles, we argue that fairness in multi-agent systems should also account for asymmetries among agents.

Furthermore, while prior work \cite{hughes2018inequity,McKeeGMDHL20,roesch2024selfishness,christoffersen2023get,vinitsky2023learning} considers environments with partial observability, they often assume that agents can see the chosen actions and received rewards of other agents. In particular, \cite{hughes2018inequity} and \cite{McKeeGMDHL20} assume that agents can observe the rewards of others in order to make comparisons, while \cite{roesch2024selfishness} instead uses the aggregate reward of all agents and allocates a portion of it back to each agent. This limits their applicability in realistic scenarios, where such information about others is often inaccessible (e.g. during inference). 

In this study, we make four key contributions. \textbf{First}, we introduce novel asymmetric versions of well-known sequential social dilemmas. These variants make it possible to analyze how natural differences between agents influence the dynamics of cooperation and how fairness-driven methods respond to such asymmetry. \textbf{Second}, we revise the definition of fairness in SSDs by accounting for agents’ potential reward ranges, enabling more meaningful comparisons across agents with different capabilities and better cooperation. \textbf{Third}, we propose an agent-based weighting mechanism for intrinsic social term, ensuring that agents with greater impact on the environment are held to higher standards of prosociality, thereby better addressing inherent asymmetries. \textbf{Finally}, we modify existing methods to operate effectively under partial observability by incorporating local social feedback, eliminating the need for direct access to other agents’ internal information.

Our experimental results demonstrate that existing fairness-based methods fail to adequately handle natural asymmetries in SSDs. In contrast, our proposed approach enables more equitable comparisons in the intrinsic social terms, preventing misleading incentives. Agent-based weighting further enhances adaptation to environmental asymmetries. Finally, the introduction of local social feedback allows fairness-based methods to achieve performance on par with globally informed counterparts, removing the necessity for constant access to other agents’ information.


\section{Background \& Related Work}

\textit{Social Dilemmas} play an important part in the field of multi-agent systems as they present the conflicts of defective and cooperative behaviors reflecting the trade-off between individual benefits and collective interests. Prisoner’s Dilemma \cite{tucker1950two} is a well-known example of a social dilemma that presents these core characteristics. Several studies model moral decision-making across various dilemmas and analyze the impact of social norms on the matrix form of the dilemmas \cite{SmitS24,TennantHM23}. 

While most studies primarily assume that agents are identical in terms of their potential actions and the rewards associated with cooperative or defection strategies, there are few studies deviating from these assumptions and exploring scenarios like the asymmetric Prisoner’s Dilemma \cite{sheposh1973asymmetry,beckenkamp2007cooperation,charness2007endogenous,andreoni1999preplay}. In asymmetric setups, various agent types emerge, where a subset of agents may experience a stronger dilemma compared to others. Under such a stronger dilemma, the advantage from defecting against a cooperator or the prospect of being exploited can be larger for a subset of agents. An advantaged agent may get more out of mutual cooperation, or a disadvantaged agent might need to pay an additional cost for cooperation \cite{zhang2022effects}. Such asymmetries exist in real life \cite{wang2021effect}. Further studies examine the impact of asymmetry in social dilemmas, especially in Prisoner's Dilemma \cite{sheposh1973asymmetry,beckenkamp2007cooperation,charness2007endogenous,andreoni1999preplay}. They show that asymmetry causes a decrease in the rate of cooperation and makes cooperation difficult to sustain unless it is actively addressed. This can be due to the propensity of disadvantaged agents to defect more frequently compared to their advantaged counterparts \cite{beckenkamp2007cooperation,janssen2012evolution,sheposh1973asymmetry,andreoni1999preplay}. It has been reported that fairness is a requirement for cooperative behavior but its definition varies between individuals, especially in asymmetric cases \cite{janssen2012evolution}. Lopez et al. \cite{lopez2022exploration} and Zhang et al. \cite{zhang2022effects} both underscore how asymmetry reduces cooperation, with lower cooperative behavior observed in groups with unequal rewards. Lopez et al. note that high-reward agents are particularly less likely to cooperate, while Zhang et al. find a significantly higher cooperation rate in symmetric reward groups.

Leibo et al. \cite{leibo2017multi} present \textit{Sequential Social Dilemmas} (SSDs), a temporal extension of a social dilemma. For examining such problems, four metrics such as efficiency, equality, sustainability and peace are introduced \cite{perolat2017multi} which respectively measure: (i) the total reward per step, (ii) how rewards are distributed among agents, (iii) the extent to which positive rewards are sustained throughout an episode, and (iv) the degree to which agent policies exhibit defecting behavior. With rational agents and without social incentives, a phenomenon called the \textit{tragedy of the commons} occurs where low sustainability is observed. Köster et al. \cite{koster2022spurious} further identify start-up and free-rider problems in a setting where agents have individualistic preferences.

Solutions to avoid the tragedy and improve group performance in SSDs consider prosocial agents that take selfless actions. There are studies that propose a parameter to control an agent's prosociality \cite{PeysakhovichL18}, an adjustment to selfishness level to induce cooperation \cite{roesch2024selfishness}, intrinsic drive based on other agent's models to calculate social impact \cite{jaques2019social} and evolving intrinsic motivation for social preferences \cite{eccles2019learning}. Alternatively, several studies allow agents to give rewards to others with the aim of incentivizing cooperation \cite{lupu2020gifting,yang2020learning}. Finally, recent approaches have explored mechanisms that go beyond intrinsic motivation. Christoffersen et al.\cite{christoffersen2023get} introduces formal contracting, where agents exchange rewards based on whether their actions are deemed cooperative or defective under a predefined contract. Another approach \cite{vinitsky2023learning} encourages agents to learn to disapprove of actions in contexts where disapproval is likely to come from others, promoting socially aligned behavior through learned social feedback.

\begin{table}[t]
    \centering
    \caption{The matrix form of a social dilemma game between the row agent $i$ and the column agent $j$. }
    \label{tab:asymmetric_social_dilemma_matrix}
    \begin{tabular}{c|c|c}
        & \textcolor{red}{C} & \textcolor{red}{D} \\ \hline 
        \textcolor{blue}{C} & $\color{blue}{R_i}, \color{red}{R_j}$ & $\color{blue}{S_i}, \color{red}{T_j}$ \\ \hline 
        \textcolor{blue}{D} & $\color{blue}{T_i}, \color{red}{S_j}$ & $\color{blue}{P_i}, \color{red}{P_j}$\\
    \end{tabular}
\end{table}

These proposed solutions often depend on global access to agents' rewards or actions, which poses a significant limitation. For fairness, agents are typically motivated to act prosocially by comparing their rewards or actions with those of others \cite{hughes2018inequity,McKeeGMDHL20,roesch2024selfishness}, requiring centralized information. Similarly, classifying behaviors as cooperative or defective often requires a centralized critic \cite{christoffersen2023get,vinitsky2023learning}. These approaches conflict with the assumption of independent learning under partial observability. Furthermore, potential asymmetry in these social dilemmas is rarely addressed in the literature, where agents have different action spaces or reward functions.


\begin{figure*}[ht]
    \centering
    \begin{subfigure}[b]{0.22\textwidth}
        \centering
        \resizebox{\textwidth}{!}{
        \begin{tikzpicture}
        
        \begin{axis}[
            title={},
            xlabel={number of other agents cooperating},
            ylabel={reward},
            xtick={0,1},
            ytick={-9,-6,-3,-2,4,5,12,15},
            ymin=-10, ymax=16,
            xmin=-0.5, xmax=1.5,
            legend style={
                at={(0.5,1.05)},
                anchor=south,
                legend columns=2,
                /tikz/every even column/.append style={column sep=0.5cm}
            },
            width=8cm,
            height=6cm,
        ]
        
        \addplot[
            color=blue,
            mark=*,
            thick
        ] coordinates {
            (0,-3)
            (1,4)
        };
        \addlegendentry{$i$ cooperate}
    
        \addplot[
            color=blue,
            mark=*,
            dashed
        ] coordinates {
            (0,-2)
            (1,5)
        };
        \addlegendentry{$i$ defect}    
        
        \addplot[
            color=red,
            mark=*,
            thick
        ] coordinates {
            (0,-9)
            (1,12)
        };
        \addlegendentry{$j$ cooperate}
    
        \addplot[
            color=red,
            mark=*,
            dashed
        ] coordinates {
            (0,-6)
            (1,15)
        };
        \addlegendentry{$j$ defect}
        
        \end{axis}
        
        \node at (4.5, 0.8) {
            \begin{tabular}{c|c|c}
                 & \textcolor{red}{C} & \textcolor{red}{D} \\ \hline 
                \textcolor{blue}{C} & \textcolor{blue}{4}, \textcolor{red}{12} & \textcolor{blue}{-3}, \textcolor{red}{15} \\ \hline 
                \textcolor{blue}{D} & \textcolor{blue}{5}, \textcolor{red}{-9} & \textcolor{blue}{-2}, \textcolor{red}{-6}\\
            \end{tabular}
        };
        
        \end{tikzpicture}
        }
        \caption{Sheposh et al. \cite{sheposh1973asymmetry}.}
        \label{fig:ex_asp_sheposh}
    \end{subfigure} 
    \quad 
    \begin{subfigure}[b]{0.22\textwidth}
        \centering
        \resizebox{\textwidth}{!}{
        \begin{tikzpicture}
        
        \begin{axis}[
            title={},
            xlabel={number of other agents cooperating},
            ylabel={reward},
            xtick={0,1},
            ytick={0,4,6,8,12,18},
            ymin=-1, ymax=19,
            xmin=-0.5, xmax=1.5,
            legend style={
                at={(0.5,1.05)},
                anchor=south,
                legend columns=2,
                /tikz/every even column/.append style={column sep=0.5cm}
            },
            width=8cm,
            height=6cm,
        ]
        
        \addplot[
            color=blue,
            mark=*,
            thick
        ] coordinates {
            (0,0)
            (1,12)
        };
        \addlegendentry{$i$ cooperate}
    
        \addplot[
            color=blue,
            mark=*,
            dashed
        ] coordinates {
            (0,6)
            (1,18)
        };
        \addlegendentry{$i$ defect}    
        
        \addplot[
            color=red,
            mark=*,
            thick
        ] coordinates {
            (0,0)
            (1,8)
        };
        \addlegendentry{$j$ cooperate}
    
        \addplot[
            color=red,
            mark=*,
            dashed
        ] coordinates {
            (0,4)
            (1,12)
        };
        \addlegendentry{$j$ defect}
        
        \end{axis}
        
        \node at (4.8, 0.8) {
            \begin{tabular}{c|c|c}
                 & \textcolor{red}{C} & \textcolor{red}{D} \\ \hline 
                \textcolor{blue}{C} & \textcolor{blue}{12}, \textcolor{red}{8} & \textcolor{blue}{0}, \textcolor{red}{12} \\ \hline 
                \textcolor{blue}{D} & \textcolor{blue}{18}, \textcolor{red}{0} & \textcolor{blue}{6}, \textcolor{red}{4}\\
            \end{tabular}
        };
        
        \end{tikzpicture}
        }
        \caption{Beckenkamp et al. \cite{beckenkamp2007cooperation}.}
        \label{fig:ex_asp_benkenkamp}
    \end{subfigure}
    \quad 
    \begin{subfigure}[b]{0.22\textwidth}
        \centering
        \resizebox{\textwidth}{!}{
        \begin{tikzpicture}
        
        \begin{axis}[
            title={},
            xlabel={number of other agents cooperating},
            ylabel={reward},
            xtick={0,1},
            ytick={2,6,7,10,13,15},
            ymin=1, ymax=16,
            xmin=-0.5, xmax=1.5,
            legend style={
                at={(0.5,1.05)},
                anchor=south,
                legend columns=2,
                /tikz/every even column/.append style={column sep=0.5cm}
            },
            width=8cm,
            height=6cm,
        ]
        
        \addplot[
            color=blue,
            mark=*,
            thick
        ] coordinates {
            (0,2)
            (1,10)
        };
        \addlegendentry{$i$ cooperate}
    
        \addplot[
            color=blue,
            mark=*,
            dashed
        ] coordinates {
            (0,7)
            (1,13)
        };
        \addlegendentry{$i$ defect}    
        
        \addplot[
            color=red,
            mark=*,
            thick
        ] coordinates {
            (0,2)
            (1,13)
        };
        \addlegendentry{$j$ cooperate}
    
        \addplot[
            color=red,
            mark=*,
            dashed
        ] coordinates {
            (0,5)
            (1,15)
        };
        \addlegendentry{$j$ defect}
        
        \end{axis}
        
        \node at (4.8, 0.8) {
            \begin{tabular}{c|c|c}
                 & \textcolor{red}{C} & \textcolor{red}{D} \\ \hline 
                \textcolor{blue}{C} & \textcolor{blue}{10}, \textcolor{red}{13} & \textcolor{blue}{2}, \textcolor{red}{15} \\ \hline 
                \textcolor{blue}{D} & \textcolor{blue}{13}, \textcolor{red}{2} & \textcolor{blue}{7}, \textcolor{red}{6}\\
            \end{tabular}
        };
        
        \end{tikzpicture}
        }
        \caption{Charnes et al. \cite{charness2007endogenous}.}
        \label{fig:ex_asp_charness}
    \end{subfigure}    
    \quad 
    \begin{subfigure}[b]{0.22\textwidth}
        \centering
        \resizebox{\textwidth}{!}{
        \begin{tikzpicture}
        
        \begin{axis}[
            title={},
            xlabel={number of other agents cooperating},
            ylabel={reward},
            xtick={0,1},
            ytick={0,3,4,6,7,9,11},
            ymin=-1, ymax=12,
            xmin=-0.5, xmax=1.5,
            legend style={
                at={(0.5,1.05)},
                anchor=south,
                legend columns=2,
                /tikz/every even column/.append style={column sep=0.5cm}
            },
            width=8cm,
            height=6cm,
        ]
        
        \addplot[
            color=blue,
            mark=*,
            thick
        ] coordinates {
            (0,0)
            (1,6)
        };
        \addlegendentry{$i$ cooperate}
    
        \addplot[
            color=blue,
            mark=*,
            dashed
        ] coordinates {
            (0,3)
            (1,9)
        };
        \addlegendentry{$i$ defect}    
        
        \addplot[
            color=red,
            mark=*,
            thick
        ] coordinates {
            (0,0)
            (1,7)
        };
        \addlegendentry{$j$ cooperate}
    
        \addplot[
            color=red,
            mark=*,
            dashed
        ] coordinates {
            (0,4)
            (1,11)
        };
        \addlegendentry{$j$ defect}
        
        \end{axis}
        
        \node at (4.8, 0.8) {
            \begin{tabular}{c|c|c}
                 & \textcolor{red}{C} & \textcolor{red}{D} \\ \hline 
                \textcolor{blue}{C} & \textcolor{blue}{6}, \textcolor{red}{7} & \textcolor{blue}{0}, \textcolor{red}{11} \\ \hline 
                \textcolor{blue}{D} & \textcolor{blue}{9}, \textcolor{red}{0} & \textcolor{blue}{3}, \textcolor{red}{4}\\
            \end{tabular}
        };
        
        \end{tikzpicture}
        }
        \caption{Andreoni et al. \cite{andreoni1999preplay}.}
        \label{fig:ex_asp_andreoni}
    \end{subfigure} 
    \caption{Schelling diagrams for example asymmetric Prisoner's Dilemma games from the literature, between agent $i$ (in blue) and agent $j$ (in red). For each agent, defective strategies lead to higher rewards.}
    \label{fig:ex_asps}
    \Description{}
\end{figure*}
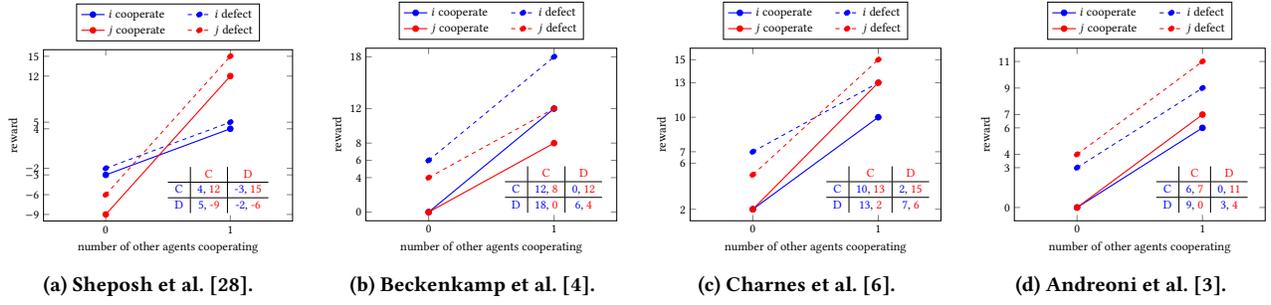

\section{Preliminaries}

Table \ref{tab:asymmetric_social_dilemma_matrix} presents a generic matrix game between agents $i$ and $j$ representing a social dilemma, where $R_i$ and $R_j$ denote the rewards when both choose to cooperate, $T_i$ and $T_j$ denote the rewards when they defect against a cooperating opponent, $S_i$ and $S_j$ denote when they cooperate while their opponent defects, $P_i$ and $P_j$ denote the rewards when both choose to defect.

A social dilemma in this form must satisfy all of the following conditions \cite{leibo2017multi,roesch2024selfishness}:
\begin{enumerate}[label={\textbf{C\arabic*:}}]
    \item $R_i > P_i$ and $R_j > P_j$: Each agent prefers mutual cooperation to mutual defection,
    
    \item $R_i > S_i$ and $R_j > S_j$: Each agent prefers mutual cooperation to being exploited by a defector,
    
    \item $2R_i > T_i + S_i$ and $2R_j > T_j + S_j$: Each agent prefers mutual cooperation to alternating between cooperation and defection,
    
    \item Either
        \begin{enumerate}[label={\textbf{C4.\alph*:}}]
            \item \textit{Greed}: $T_i > R_i$ and $T_j > R_j$: there is a temptation to defect against a cooperator,
            \item or \textit{Fear}: $P_i > S_i$ and $P_j > S_j$: defection in face of being exploited is preferred.
        \end{enumerate}
\end{enumerate}

\textit{Prisoner’s Dilemma} like social dilemmas involve both \textit{greed} and \textit{fear} factors satisfying C4.a and C4.b at the same time. If only the \textit{greed} factor is present, it is known as a \textit{Chicken Game}, and if only the \textit{fear} factor is present, it is called a \textit{Stag Hunt Game} \cite{roesch2024selfishness,marl-book}.

\begin{definition}[Sequential Social Dilemma]
A sequential social dilemma (SSD) is a tuple $(\mathcal{M}, \Pi^C, \Pi^D)$ where $\mathcal{M}$ is a partially observable stochastic game such that there exist states $s \in S$ for which the induced empirical reward matrix satisfies the social dilemma inequalities (C1-C4), and $\Pi^C$ and $\Pi^D$ are disjoint sets of policies that are said to implement cooperation and defection, respectively.
\label{def:ssd}
\end{definition}

Such a game $\mathcal{M}$ with $N$-agents is defined \cite{marl-book} as a tuple \\ $\langle \mathcal{N}, S, \{A_i\}_{i \in \mathcal{N}}, T, \{R_i\}_{i \in \mathcal{N}}, \Omega, \{O_i\}_{i \in \mathcal{N}} \rangle$ where
\begin{itemize}
    \item $\mathcal{N} = \{1, 2,...,N\}$ is the set of agents,
    \item $S$ is the finite set of states,
    \item $A$ is the joint set of actions such that $A = A_1 \times A_2 \times ... \times A_N$ where $A_i$ is the finite set of actions for agent $i$,
    \item $\mathcal{T}: S \times A \rightarrow \Delta(S)$ is the transition function with $\Delta(S)$ as the set of discrete probability distributions over $S$,
    \item $\mathcal{R}_i: S \times A \times S \rightarrow \mathbb{R}$ defines the reward function for agent $i$,
    \item $\Omega: S \times A \rightarrow \Delta(O)$ specifies the observation function
    \item $O = O_1 \times O_2 \times ... \times O_N$ is the joint set of observations so that the agent $i$ receives a local observation $o_i \in O_i$.
\end{itemize}

Agents learn a joint policy $\pi = (\pi_1, ..., \pi_N)$ where each agent $i$ aims to learn its own policy $\pi_i$ that maximizes its expected return given the policies of the others $\pi_{-i}$, $U_i(\pi_i, \pi_{-i}) = \mathbb{E}_{s^0 \sim \mu} [V_i^\pi(s^0)]$ where $V_i^\pi(s) = \mathbb{E}_{\pi, \mathcal{T}} [\sum_{t=0}^\infty \gamma^t R_i(s^t, a^t, s^{t+1}) \; | \; s^0 = s]$ and $\mu$ is the initial state distribution with $\gamma \in [0, 1]$ as the discount factor, $s^t \in S$ the state and $a^t \in A$ the joint action at time step $t$.

\subsection{Base Methods}

Inequity Aversion (IA) \cite{hughes2018inequity} and Social Value Orientation (SVO) \cite{McKeeGMDHL20} are used in this study as representative fairness-based intrinsic motivation methods to evaluate how our approach better addresses the fairness problem in asymmetric settings. Both methods provide an intrinsic drive for fairness that takes into account the temporal feature of the problem environments, calculate the temporally smoothed reward of an agent $i$ at time step $t$ as
\begin{align}
    e_{i}^t = \gamma \cdot \lambda \cdot e_{i}^{t-1} + r_{i}^t,
    \label{eq:tsr}
\end{align}
where $r_{i}^t$ is the reward taken at time step $t$ and $\lambda$ is a hyperparameter controlling the decay rate. Temporally smoothed reward of an agent is an indication of how well the agent is doing within an episode and it is a base point for comparison.

Using these smoothed rewards, IA modifies the reward of the agent $i$ at time step $t$ as 
\begin{align}
    \hat{r}_i^t = \; & r_i^t \nonumber \\
    &- \frac{\alpha}{N - 1} \cdot \sum_{j \neq i} \max(e_j^t - e_i^t, 0) \nonumber \\
    &- \frac{\beta}{N - 1} \cdot \sum_{j \neq i} \max(e_i^t - e_j^t, 0),
    \label{eq:ia}
\end{align}
where $\alpha$ and $\beta$ are parameters to control \textit{disadvantageous} and \textit{advantageous} inequity, respectively. A higher $\alpha$ penalizes the agent when it receives less reward than others, while a higher $\beta$ penalizes when it receives more.

Similarly, using the temporally smoothed rewards, SVO calculates a reward angle for agent $i$ at time step $t$ as 
\begin{align}
    \theta^t_i = \text{atan} \Big( \frac{e_{-i}^{t}}{e_i^t} \Big),
    \label{eq:svo_reward_angle}
\end{align}
where $e_{-i}^{t} = \frac{1}{N-1} \sum_{j \neq i} e_j^t$. The method requires a target social value orientation $\theta^{SVO}_i$ for an agent $i$ and modifies the reward of the agent as 
\begin{align}
    \hat{r}_i^t = \; & r_i^t - w \cdot \left|\theta^{SVO}_i - \theta^t_i\right|,
    \label{eq:svo}
\end{align}
with $w$ as a weight to control the effect of SVO. Having $\theta^{SVO} = 45\degree$ creates an incentive to be equal with the rest of the group in terms of rewards.

\subsection{Asymmetry in Social Dilemmas}

Social dilemmas can feature various asymmetries between agents. These asymmetries may arise from differences in available actions (abilities) or rewards (payoffs). Our analysis focuses on asymmetries that can be reflected in rewards and potentially alter agent behaviors.

We define asymmetry in a social dilemma as a situation in which agents differ in how they experience the outcomes of key strategic interactions. Specifically, a social dilemma is \textit{asymmetric} if there is any inequality in the reward values assigned to each agent: mutual cooperation yields different rewards (\( R_i \neq R_j \)); the temptation to defect against a cooperator varies between agents (\( T_i \neq T_j \)); the cost of being exploited while cooperating differs (\( S_i \neq S_j \)); or mutual defection results in unequal rewards (\( P_i \neq P_j \)). Although all agents face the same social dilemma, differences in rewards under identical situations create unequal incentives, which can lead to different experiences and behaviors within the same environment.

\figref \ref{fig:ex_asps} presents several examples of asymmetric Prisoner's Dilem\-ma games from the literature. In all cases, the defining inequalities of the social dilemma are preserved. That is, in each case, defection yields a higher reward than cooperation for any individual agent. However, the introduced asymmetry leads to different characteristics across the examples. While asymmetry is present in all cases of the games, the advantage shifts depending on the context.

In \figref \ref{fig:ex_asp_benkenkamp} and \figref \ref{fig:ex_asp_andreoni}, one agent consistently receives equal or greater rewards than the other, regardless of the outcome. In contrast, \figref \ref{fig:ex_asp_sheposh} shows a more nuanced dynamic: agent $i$ (blue) earns more than agent $j$ (red) when facing a defecting opponent, but less when the opponent cooperates. A similar role reversal appears in \figref \ref{fig:ex_asp_charness}, though it only occurs under mutual defection—agent $i$ benefits more than agent $j$ in that case, whereas the advantage flips in mutual cooperation. 

These examples illustrate that asymmetry in reward structure does not always imply a consistent advantage for a particular agent across all strategic scenarios.

Extending from matrix games to more complex problems, based on Leibo et al.'s definition \cite{leibo2017multi} of an SSD, in this general setting, we can define the rewards of an agent $i$ for a state $s$ satisfying social dilemma conditions as:
\begin{align*}
    R_i(s) &= V_i^\pi(s) \text{ when } \pi_i \in \Pi^C, \pi_{-i} \in \Pi^C,\\
    T_i(s) &= V_i^\pi(s) \text{ when } \pi_i \in \Pi^D, \pi_{-i} \in \Pi^C,\\
    S_i(s) &= V_i^\pi(s) \text{ when } \pi_i \in \Pi^C, \pi_{-i} \in \Pi^D,\\
    P_i(s) &= V_i^\pi(s) \text{ when } \pi_i \in \Pi^D, \pi_{-i} \in \Pi^D.
\end{align*}

\begin{definition}[Asymmetric Sequential Social Dilemma]
An SSD becomes \textit{asymmetric} if and only if there exist agents $i, j \in \mathcal{N}$ with $i \neq j$ and a state $s \in S$ such that at least one of $R_i(s) \neq R_j(s)$, $T_i(s) \neq T_j(s)$, $S_i(s) \neq S_j(s)$, or $P_i(s) \neq P_j(s)$ holds and the terms satisfy the conditions C1 to C4.     
\end{definition}


\section{Proposed Approach}




Existing methods like IA and SVO promote cooperation by comparing agents' rewards (often coupling intrinsic and extrinsic signals in a single update \cite{schaefer2022derl}), assuming that differences reflect variations in cooperative behavior—as in symmetric games such as the Prisoner's Dilemma. In asymmetric settings, however, identical policies can yield different rewards due to inherent differences in roles or capabilities. Equality-driven methods may misinterpret these natural differences as unfairness, penalizing agents who are fully cooperating (Eqs. \ref{eq:ia} and \ref{eq:svo}). As a result, such methods can even incentivize harmful behavior, as lower-reward agents may defect to reduce the gap, undermining both cooperation and fairness. The key issue is that raw reward comparisons ignore underlying asymmetries.

\subsection{Adjusting for Agent Potential}

To fairly evaluate cooperation or defection, we propose that agents must be assessed relative to their own reward scales. That is, an agent’s position in its own outcome range, between its lowest and highest possible rewards ($S_i < P_i < R_i < T_i$), should be considered. Since this range can differ across agents, comparing their raw rewards directly is misleading. Instead, each agent's behavior should be located within its own cooperative-defective range before any cross-agent comparison is made.

\begin{table}[t]
    \centering
    \caption{Example asymmetric Prisoner's Dilemma game between the row agent $i$ (blue) and the column agent $j$ (red) before (a) and after normalization (b).}
    \begin{tabular}{cc}
        \begin{minipage}{0.45\linewidth}
            \centering
            \caption*{(a) original}
            \begin{tabular}{c|c|c}
                & \textcolor{red}{C} & \textcolor{red}{D} \\ \hline 
                \textcolor{blue}{C} & $\color{blue}{-1}, \color{red}{-6}$ & $\color{blue}{-3}, \color{red}{-5}$ \\ \hline 
                \textcolor{blue}{D} & $\color{blue}{0}, \color{red}{-8}$ & $\color{blue}{-2}, \color{red}{-7}$\\
            \end{tabular}
        \end{minipage}
        &
        \begin{minipage}{0.45\linewidth}
            \centering
            \caption*{(b) after normalization}
            \begin{tabular}{c|c|c}
                & \textcolor{red}{C} & \textcolor{red}{D} \\ \hline 
                \textcolor{blue}{C} & $\color{blue}{0.67}, \color{red}{0.67}$ & $\color{blue}{0}, \color{red}{1}$ \\ \hline 
                \textcolor{blue}{D} & $\color{blue}{1}, \color{red}{0}$ & $\color{blue}{0.33}, \color{red}{0.33}$\\
            \end{tabular}
        \end{minipage}
    \end{tabular}
    
    \label{tab:ex_asymmetric_social_dilemma}
\end{table}

Consider the asymmetric Prisoner's Dilemma shown in Table \ref{tab:ex_asymmetric_social_dilemma}, where the rewards represent the number of years each prisoner is sentenced. In this scenario, the agent $j$ (red) has a prior conviction, resulting in an additional 5 years added to its sentence for any outcome. Despite the asymmetry, the social dilemma conditions C1 to C4 are still satisfied for both agents: that is, $S_i < P_i < R_i < T_i$ and $S_j < P_j < R_j < T_j$. However, in this example, the temptation reward $T_j$ of agent $j$ (red) when defecting against a cooperating agent is numerically smaller than the sucker reward $S_i$ of agent $i$ (blue) in the same case. This is allowed under the dilemma inequalities but can lead to misleading interpretations. A direct comparison between $T_j$ and $S_i$ might suggest that agent $j$ is more cooperative than agent $i$ even though agent $j$ is defecting and agent $i$ is cooperating. To address this issue, a normalization step is applied, resulting in the transformed game shown in Table \ref{tab:ex_asymmetric_social_dilemma}. In the normalized version, each agent's behavior is expressed relative to their own reward range, enabling a more accurate comparison of the relative advantages and disadvantages between agents.

As a general approach, we propose a normalization step to adjust agents' rewards over time in an SSD. Both IA and SVO utilize temporally smoothed rewards as a measure of how well one agent is doing at a time step. A \textit{normalized} temporally smoothed reward is defined as
\begin{align}
    \hat{e}^t_i = \frac{e_i^t - e_i^{min}}{e_i^{max} - e_i^{min}},
    \label{eq:tsr_norm}
\end{align}
where $e^t_i$ is the temporally smoothed reward as defined in Eq. \ref{eq:tsr} and $e_i^{max} = \max_t(e_i^0, ..., e_i^t)$ and $e_i^{min} = \min_t(e_i^0, ..., e_i^t)$. With this modification, IA and SVO become fairer and better suited to asymmetric settings, using the normalized temporally smoothed rewards in Eqs. \ref{eq:ia} and \ref{eq:svo_reward_angle}. 

This modification provides several improvements. First, it reduces the unintended incentive to punish cooperative agents who happen to earn higher rewards, helping to maintain the focus on comparing cooperative versus defective policies. As a result, agents are evaluated based on how well they perform relative to their own potential. 
Second, the normalization step ensures that values remain bounded and non-negative, providing greater control for IA and preventing negative reward angles in SVO. In environments with negative rewards, SVO can produce negative angles, which may be ambiguous. A negative value can arise in two distinct scenarios: (1) agent $i$ receives a positive reward while others receive negative rewards, or (2) agent $i$ receives a negative reward while others receive positive rewards. In the first case, agent $i$ is acting selfishly; in the second, it is being highly altruistic. Despite reflecting opposite behaviors, both cases may yield the same negative angle in SVO. Normalization eliminates such ambiguity by preventing negative values, leading to more interpretable and consistent measurements.

\begin{table*}[t]
\centering
\caption{Agent types used in the \texttt{Coins} and \texttt{Harvest}. In \texttt{Coins}, ``mismatch coin'' refers to collecting the other agent's coin.}
\label{tab:agent_types}
\begin{tabular}{|l|c|l|c|}
\hline
\textbf{Agent Type} & \textbf{Reward per Item} & \textbf{Abilities / Effects} & \textbf{Environment(s)} \\
\hline
Standard      & +1.0  & Movement; zap in \texttt{Harvest} only & \texttt{Harvest}, \texttt{Coins} \\
Low-reward & +0.5  & Movement; in \texttt{Coins}, collecting a mismatch coin penalizes the other by $-3.0$ & \texttt{Harvest}, \texttt{Coins} \\
High-reward & +1.5  & Movement; in \texttt{Coins}, collecting a mismatch coin penalizes the other by $-1.0$ & \texttt{Harvest}, \texttt{Coins} \\
Wide-zap      & +1.0  & Larger zapping radius & \texttt{Harvest} \\
Spawn-biased & +1.0  & Collecting a mismatch coin makes only this agent's coins spawn for 25 steps & \texttt{Coins} \\
\hline
\end{tabular}
\end{table*}

\subsection{Agent-Based Weighting of Social Influence}

To better address asymmetry, differences in agents’ potential impact on the environment should be reflected in their commitment to social preferences. Agents with inherent advantages must be held to a stricter standard of social responsibility. This can be implemented using agent-specific weighting by introducing a term, the \textit{social drive modifier} as we call it, into the Eqs. \ref{eq:ia} and \ref{eq:svo}.

Hence, we extend the definition for the weights of disadvantageous and advantageous inequity for agent $i$ in IA as follows: 
\begin{align}
    \bar{\alpha}_i = \phi_i \cdot \alpha , \nonumber \\
    \bar{\beta}_i = \phi_i \cdot \beta, 
\end{align}

where $\phi_i \geq 0$ is the social drive modifier of agent $i$. Similarly, in SVO, the weight of agent $i$ is redefined as 
\begin{align}
    \bar{w}_i = \phi_i \cdot w.
\end{align}
 
With these adjustments, an agent with a larger potential of influencing the environment (same action causing a bigger effect) can have a higher intrinsic drive to be prosocial. While the original parameters such as $\alpha, \beta$ and $w$ in IA and SVO determine the importance of the social term in the reward functions in general, $\phi$ allows agent-specific adjustment to address inherit differences between agents.

\subsection{Localizing Social Feedback}

Finally, we propose a decentralized approach in order to make IA and SVO consistent with the partially observable nature of these settings as defined in Def. \ref{def:ssd}. In this approach, each agent maintains local estimates of other agents’ temporally smoothed rewards, which are updated through direct communication when available. 

Let $\hat{e}_{i,j}^t$ be agent $i$'s estimate at time $t$ of agent $j$'s normalized temporally smoothed reward. Let $\tau_{i,j}^t$ denote the time step at which $\hat{e}_{i,j}^t$ was last updated, and let $\mathcal{N}_i^t$ be the set of other agents visible to agent $i$ at time step $t$. After each time step $t$, the temporally smoothed rewards and agents' local estimates are updated according to Alg. \ref{alg:tsr_update}. After these steps, each agent $i$ uses its own estimates for other agents in the Eqs. \ref{eq:ia} and \ref{eq:svo_reward_angle} where $e_j^t$ is replaced by $e_{i, j}^t$.

\begin{algorithm}[t]
\caption{Update on Temporally Smoothed Rewards}
\label{alg:tsr_update}
\begin{algorithmic}[1]
    \For{each agent $i$} 
        \LineComment{Update estimates for non-visible agents}
        \For{each agent $k \notin \mathcal{N}_i^t$}
            \State $j^\ast \gets \arg\max_{j \in \mathcal{N}_i^t} \tau_{j,k}^{t-1}$
            \State $\hat{e}_{i,k}^t \gets \hat{e}_{j^\ast,k}^{t-1}$
            \State $\tau_{i,k}^t \gets \tau_{j^\ast,k}^{t-1}$
        \EndFor

        \LineComment{Update own temporally smoothed reward}
        \State $e_{i}^t \gets \gamma \cdot \lambda \cdot e_{i}^{t-1} + r_{i}^t$
        \State $e_i^{max} \gets \max_t(e_i^0, ..., e_i^t)$
        \State $e_i^{min} \gets \min_t(e_i^0, ..., e_i^t)$
        \State $\hat{e}^t_i \gets \frac{e_i^t - e_i^{min}}{e_i^{max} - e_i^{min}}$

        \LineComment{Update estimates for visible agents}
        \For{each agent $j \in \mathcal{N}_i^t$}
            \State $e_{i,j}^t \gets e_j^t$
            \State $\tau_{i,j}^t \gets t$
        \EndFor
    \EndFor
\end{algorithmic}
\end{algorithm}

This localized mechanism relaxes the strong assumption made by methods like IA and SVO, which typically require full access to all agents' temporally smoothed rewards at every time step. By adjusting to partial observability, these methods are adapted to more realistic settings where agents have limited local information. In this setup, an agent can directly observe how well a visible agent is performing and, through communication, indirectly obtain information about how non-visible agents were performing in the past via the estimates held by other visible agents.

These three modifications make the Fair\&Local versions of IA and SVO better suited to asymmetric SSDs under partial observability. As raw equality can incentivize defection, the objective is fairness through mutual cooperation, allowing agents to achieve the best outcomes within their respective ranges.
With this approach, agents learn cooperative policies more effectively without requiring global information.


\section{Experiments}

Our experiments study the following questions:
\begin{enumerate}
\item How does asymmetry in sequential social dilemmas affect learning, both with and without equality-based intrinsic motivation?
\item Can accounting for agent potential and applying agent-based weighting mitigate the challenges introduced by asymmetry in terms of cooperation?
\item Can methods that rely on local social feedback match the performance of their global-access counterparts?
\end{enumerate}

\subsection{Environments}

\begin{figure}[h]
    \centering
    \begin{subfigure}[t]{0.3\columnwidth}
        \centering
        \includegraphics[height=0.1\textheight]{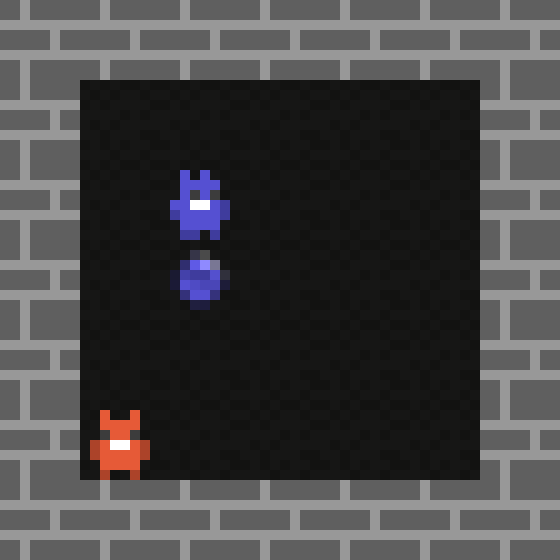}
        \caption{\texttt{Coins}}
        \label{fig:coins_env}
    \end{subfigure}
    \hfill
    \begin{subfigure}[t]{0.6\columnwidth}
        \centering
        \includegraphics[height=0.1\textheight]{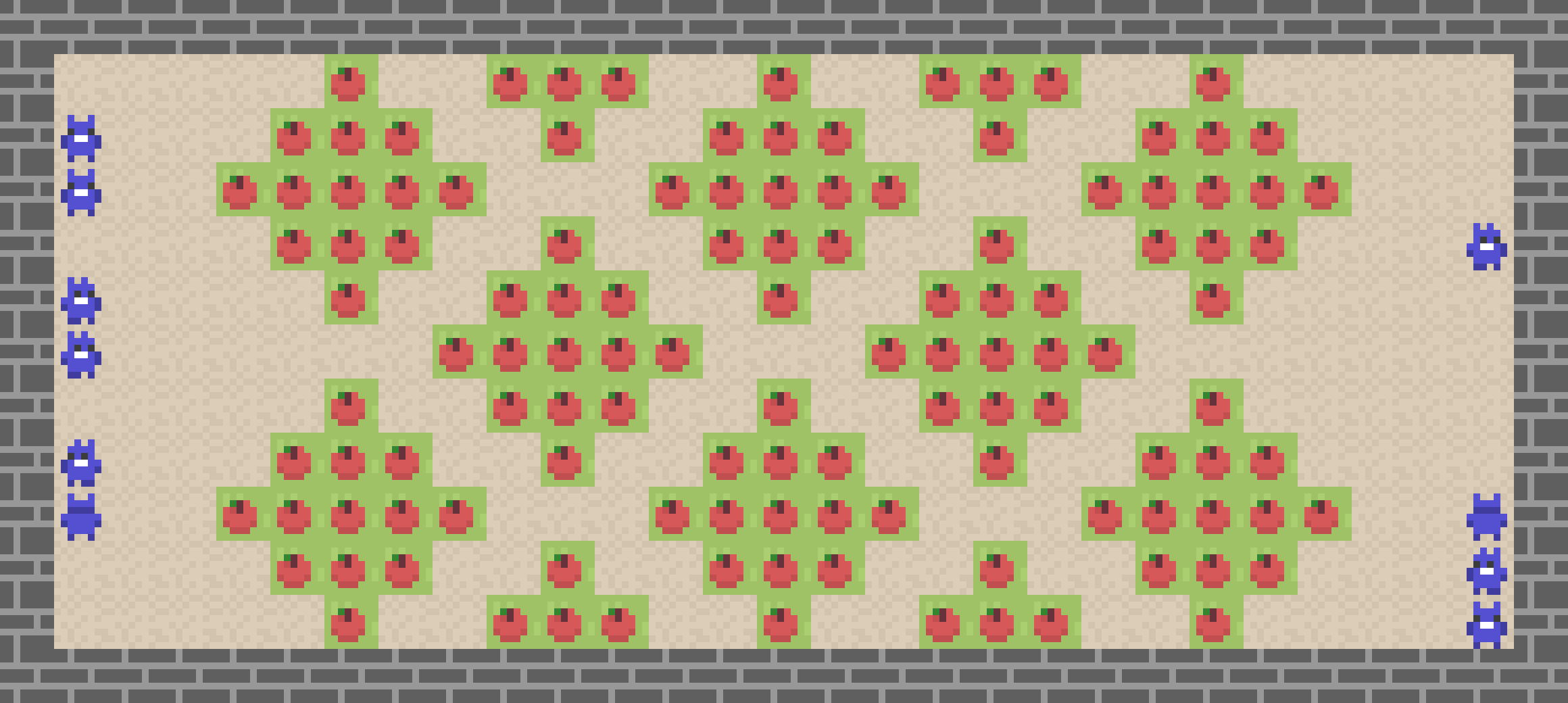}
        \caption{\texttt{Harvest}}
        \label{fig:harvest_env}
    \end{subfigure}

    \caption{(a) \texttt{Coins} environment with 2 standard (1 red and 1 blue) agents; 
             (b) \texttt{Harvest} environment with 10 standard agents.}
    \label{fig:harvest_and_coins_environments}
    \Description{}
\end{figure} 

We experiment using two well-known benchmark environments\footnote{The link to the code will be available in camera-ready.}.

The \texttt{Coins} environment \cite{lerer2017maintaining,eccles2019learning}, shown in \figref \ref{fig:coins_env}, involves two agents, red and blue. Coins of both colors spawn at a given rate. Collecting any coin yields +1 reward, but picking up the other agent’s coin penalizes that agent by –2. A cooperative agent must therefore avoid collecting mismatch coins.

The \texttt{Harvest} environment \cite{perolat2017multi,PeysakhovichL18}, shown in \figref \ref{fig:harvest_env}, features 10 agents collecting apples, each worth +1. Apple regrowth depends on leaving nearby apples uncollected; over-harvesting depletes resources. Agents can also zap others with a beam, temporarily removing them from the environment for 25 steps. The dilemma arises because quickly collecting apples maximizes individual reward but undermines long-term group benefit.

Unlike matrix-form social dilemmas, SSDs do not have fixed cooperate or defect actions; instead, these labels must be assigned to behaviors. In \texttt{Harvest}, cooperation means avoiding apple collection in low-density areas, while in \texttt{Coins} it means collecting only one’s own coins. In \texttt{Harvest}, zapping non-defective agents or harvesting the last remaining apples are considered defective behaviors. 

To study the role of asymmetry, we use variations of the environments where agents are divided into two subgroups with different characteristics. Their social dilemma properties are verified using Schelling diagrams in Appendix \ref{sec:env_verification}.

\subsubsection{Asymmetry in rewards} Here, base rewards are scaled: low-reward agents receive half the original reward, while high-reward agents receive one and a half times. In \texttt{Coins}, penalties are also scaled: when a low-reward agent collects the coin of a high-reward agent, the latter incurs a penalty of –3, while the reverse case results in only –1. These multipliers were chosen to make asymmetry visible without destabilizing training. These changes yield two variants: \texttt{Coins} with one high- and one low-reward agent (\textit{asymmetry in coin rewards}), and \texttt{Harvest} with five of each (\textit{asymmetry in apple rewards}).

\subsubsection{Asymmetry in actions} Here, the mechanics of actions differ between subgroups. In \texttt{Harvest}, wide-zap agents have an extended laser radius. In \texttt{Coins}, the spawn-biased agent causes only its own coins to appear for 25 steps after collecting the other’s coin. Rewards remain unchanged; only the effects of actions differ. These yield two additional variants: \texttt{Coins} with one standard and one spawn-biased agent (\textit{asymmetry in coin spawn}), and \texttt{Harvest} with five standard and five wide-zap agents (\textit{asymmetry in zap radius}).

Table \ref{tab:agent_types} summarizes the agent types used in our experiments.

\begin{figure*}
    \centering
    \begin{subfigure}[b]{0.48\textwidth}
        \centering
        \includegraphics[width=0.38\textheight]{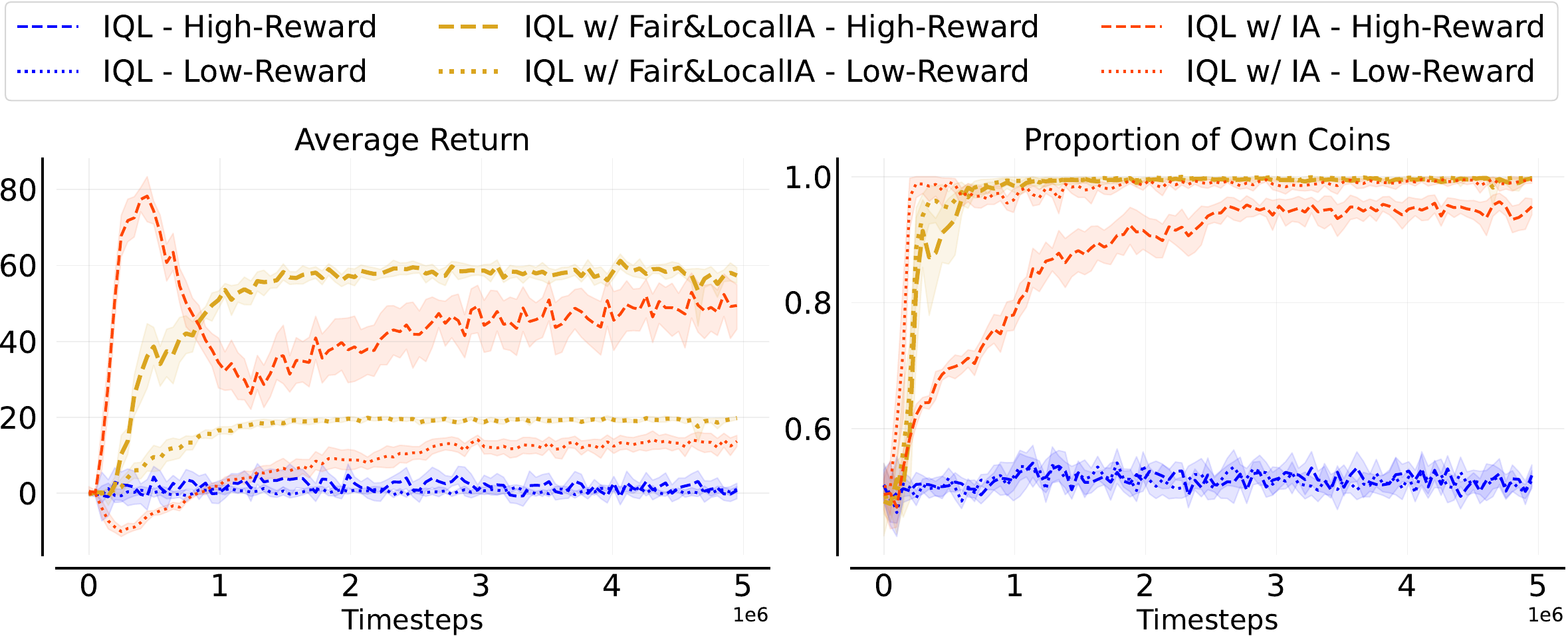}
    \end{subfigure} 
    \quad
    \begin{subfigure}[b]{0.48\textwidth}
        \centering
        \includegraphics[width=0.38\textheight]{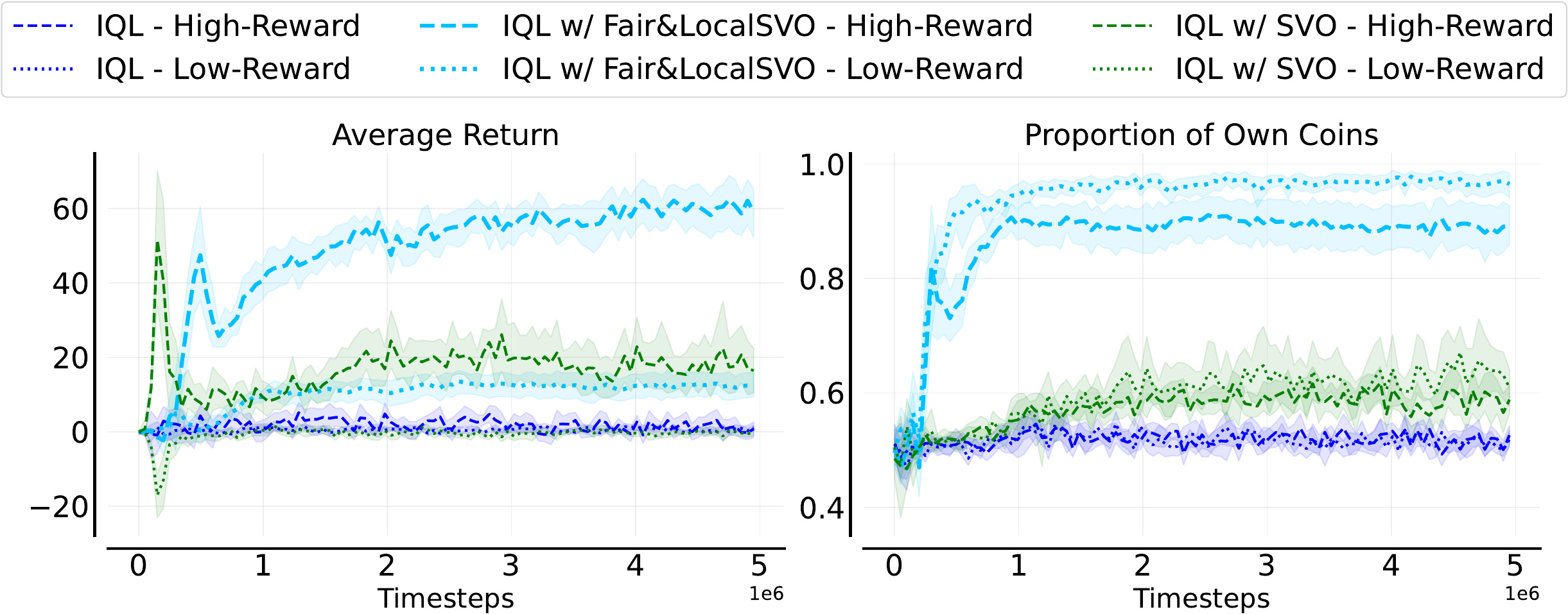}
    \end{subfigure}    
    \caption{Performance of IA, SVO, and their Fair\&Local versions in \texttt{Coins} with \textit{asymmetry in coin rewards}. Fair\&LocalIA and Fair\&LocalSVO shows faster and more stable convergence to prosocial behavior with higher returns than their counterparts.}
    \label{fig:asymmetric_coins__rf_1adv_VS_1dis_vs_fair_local}
    \Description{}
\end{figure*} 

\begin{figure*}
    \centering
    \begin{subfigure}[b]{1\textwidth}
        \centering
        \includegraphics[width=0.8\textheight]{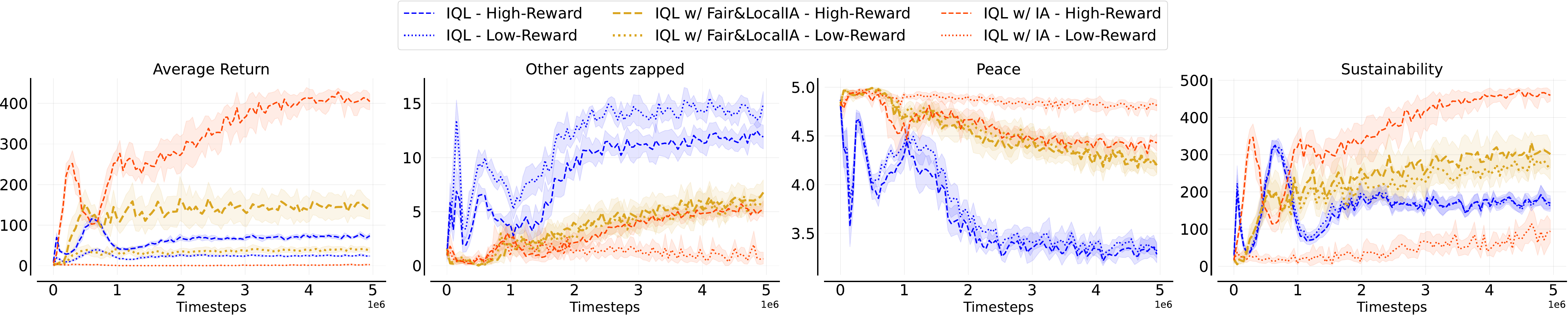}
    \end{subfigure}  
    \\
    \begin{subfigure}[b]{1\textwidth}
        \centering
        \includegraphics[width=0.8\textheight]{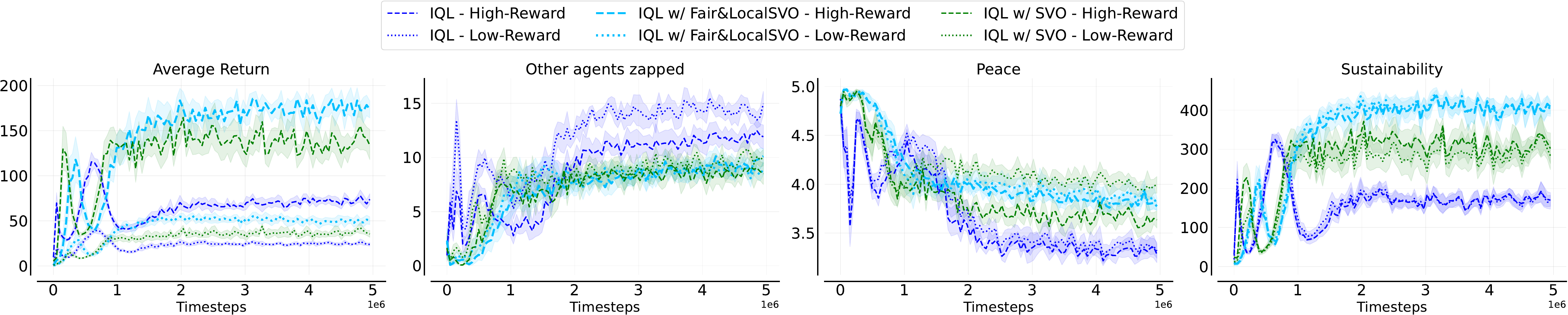}
    \end{subfigure} 
    \caption{Performance of IA, SVO, and their Fair\&Local versions in \texttt{Harvest} environment with \textit{asymmetry in apple rewards}. Fair\&Local versions lead to higher returns and better sustainability by eliminating the gap in peace between agent types.}
    \label{fig:asymmetric_commons_harvest__rf_5adv_VS_5dis_vs_fair_local}
    \Description{}
\end{figure*}

\subsection{Baselines}






Agents are trained with Independent DQN (IQL) \cite{papoudakis2021benchmarking}. We use three baselines: plain \textit{IQL} to show the direct effect of asymmetry, \textit{IQL with IA} (Eq. \ref{eq:ia}) to test inequity aversion’s limits, and \textit{IQL with SVO} (Eq. \ref{eq:svo}) to assess social value orientation’s ability to handle asymmetry. Our proposed methods, Fair\&\-LocalIA and Fair\&\-LocalSVO, extend these approaches with three modifications, enabling fairer adaptation to asymmetric environments. The implementation details and used hyperparameters are given in Appendix \ref{sec:implementation}.

\begin{figure*}
    \centering
    \begin{subfigure}[b]{0.48\textwidth}
        \centering
        \includegraphics[width=0.38\textheight]{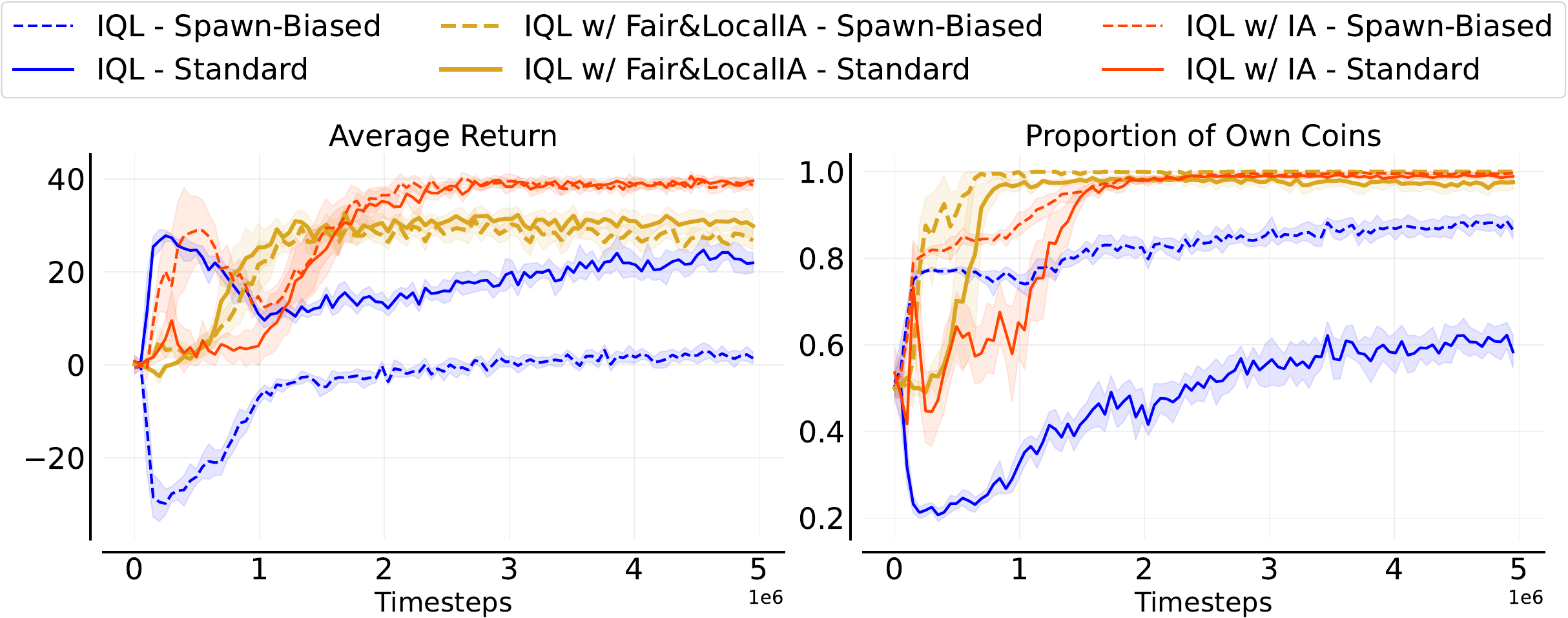}
    \end{subfigure} 
    \quad
    \begin{subfigure}[b]{0.48\textwidth}
        \centering
        \includegraphics[width=0.38\textheight]{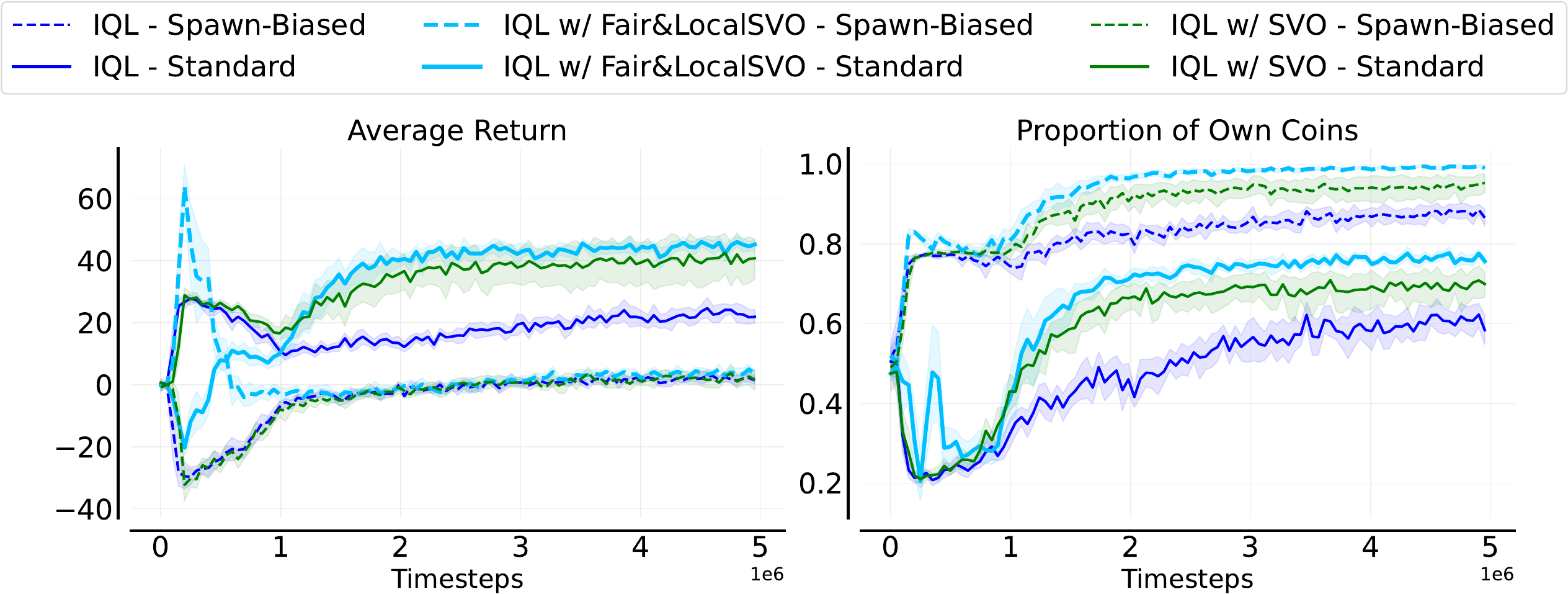}
    \end{subfigure}    
    \caption{Performance of IA, SVO, and their Fair\&Local versions in \texttt{Coins} environment with \textit{asymmetry in coin spawn}. The proposed approach shows faster convergence to more cooperative policies.}
    \label{fig:asymmetric_coins__as_1adv_VS_1dis_vs_fair_local}
    \Description{}
\end{figure*} 

\begin{figure*}
    \centering
    \begin{subfigure}[b]{1\textwidth}
        \centering
        \includegraphics[width=0.8\textheight]{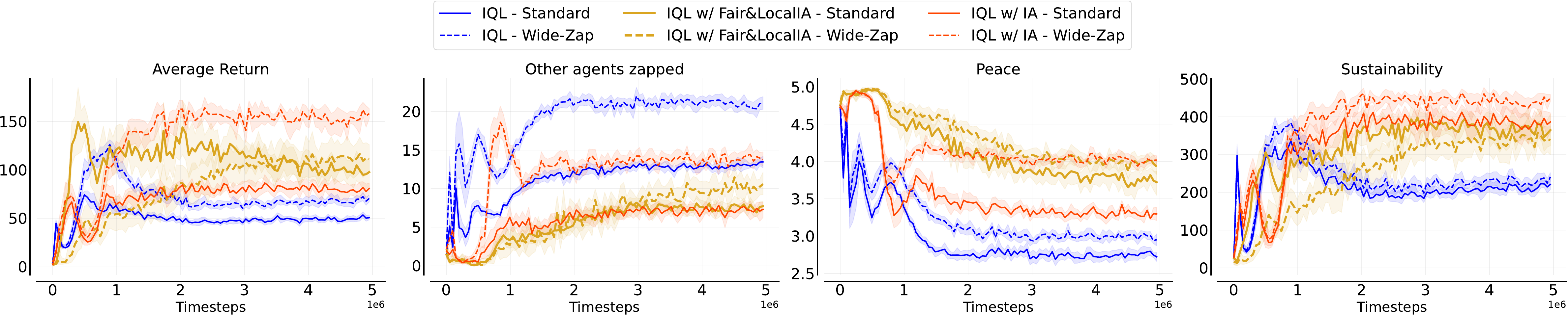}
    \end{subfigure} 
    \\
    \begin{subfigure}[b]{1\textwidth}
        \centering
        \includegraphics[width=0.8\textheight]{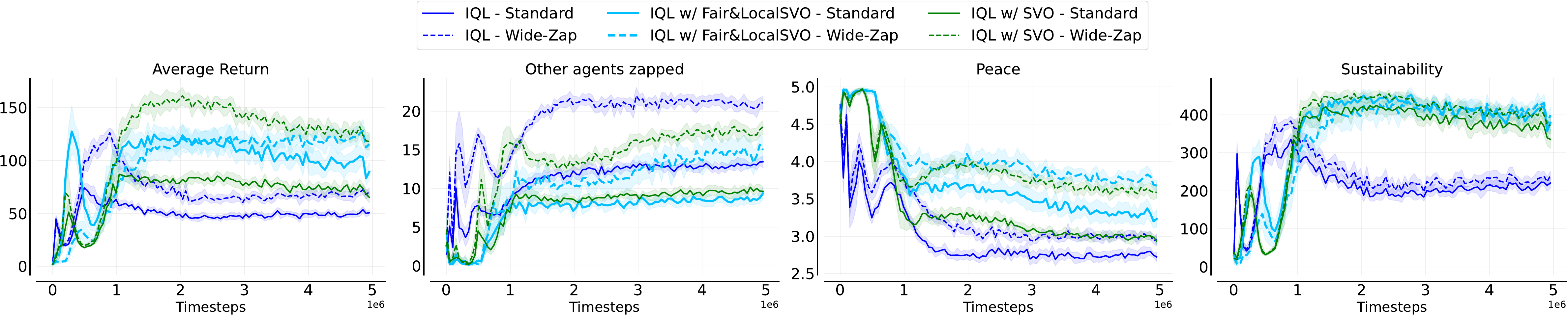}
    \end{subfigure} 
    \caption{Performance of IA, SVO, and their Fair\&Local versions in \texttt{Harvest} with \textit{asymmetry in zapping}. Fair\&Local versions with different weights for different types of agents, lead to more cooperative policies with higher peace and similar returns for each agent type.}
    \label{fig:asymmetric_commons_harvest__as_5adv_VS_5dis_vs_fair_local}
    \Description{}
\end{figure*}

\subsection{Evaluation Metrics}
The experiments are evaluated using several metrics that capture different aspects of agent behavior.

Let $G_i = \sum_{t = 0}^T r_i^t$ denote the return of agent $i$ in an episode of $T$ time steps, where $r_i^t$ is the reward received by agent $i$ at time step $t$. The \textit{average return} is defined as $\hat{G} = \frac{1}{N} \sum_{i=1}^N G_i,$ where higher values indicate better overall learning performance. 

In \texttt{Coins}, the prosocial behavior can be measured by \textit{the proportion of own coins} as the ratio of number of own collected coins to the number of total collected coins, averaged over agents. 

Having the zapping action and a different dynamic, \texttt{Harvest} requires other metrics. \textit{Sustainability} measures how evenly positive rewards are distributed over time and is defined as $S = \frac{1}{N} \sum_{i=1}^N t_i,$ where $t_i = \mathbb{E}[\, t \mid r_i^t > 0 \,]$ represents the expected time step at which agent $i$ receives positive reward. Lower values suggest that most rewards were collected early in the episode. \textit{Peace} captures the level of conflict among agents and is defined as $P = N - \frac{1}{T} \sum_{i=1}^N \sum_{t=1}^T \mathbb{I}\{\text{agent } i \text{ timed out on step } t\},$ where $P \in [0, N]$, and lower values indicate more aggression in the environment. \textit{The number of other agents zapped} by agent $i$, is denoted as $Z_i$, with average $\hat{Z} = \frac{1}{N} \sum_{i=1}^N Z_i$ among agents.


The experimental results measure these metrics per agent type in order to identify the behavioral differences between agent types. 

\subsection{Results}

\subsubsection{Asymmetry in Rewards}

\figref \ref{fig:asymmetric_coins__rf_1adv_VS_1dis_vs_fair_local} shows that in \texttt{Coins} with one high-reward and one low-reward agent, independent learners without prosocial incentives fail to develop cooperative strategies, indiscriminately collecting coins and converging to zero average return. IA encourages some cooperation but struggles to handle the inherent asymmetry, especially for the agent with reward advantage, whereas Fair\&LocalIA enables both agent types to fully cooperate, collecting only their own coins and achieving faster convergence. SVO offers only limited improvement over the baseline IQL, while Fair\&LocalSVO more effectively promotes prosocial behavior, yielding higher returns as agents primarily collect their respective coins.

In \figref \ref{fig:asymmetric_commons_harvest__rf_5adv_VS_5dis_vs_fair_local}, the tragedy of commons is visible in \texttt{Harvest} with IQL, converging to mutual defection with a decrease in peace metric. It shows that the low-reward agents are more aggressive in terms of zapping. IQL with IA, gives a high punishment for low-reward agents as they fall behind of high-reward agents, causing them to stop moving altogether. As a result, high-reward agents collect all remaining apples, showing high sustainability among them and low peace since low-reward agents are not in the apple collection area. IQL with Fair\&LocalIA, allows a better adjustment to their respective reward scales, making both types of agents actively cooperate in the environment, showing higher peace with lower zapping behavior and balanced sustainability compared to IQL with IA. While SVO shows higher returns and peace compared to IA, its wrong incentive for low-reward agents is clear in peace metric. Both types of agents zap others at the same amounts but the gap in the peace metric suggests that it is a targeted action. High-reward agents have a lower peace value, meaning that low-reward agents specifically zap the high-reward ones since this reduces the gap in rewards according to Eq. \ref{eq:svo}. This wrong incentive to defect for low-reward agents is eliminated with Fair\&LocalSVO, removing the gap in peace. IQL with Fair\&LocalSVO also outperforms IQL with SVO in terms of sustainability metric.

\subsubsection{Asymmetry in Actions}

In \texttt{Coins} with \textit{asymmetry in coin spawn}, \figref \ref{fig:asymmetric_coins__as_1adv_VS_1dis_vs_fair_local} shows that both agents without any social incentives act selfishly, yet this selfish act causes spawn-biased agent to collect less returns than the standard one since its defective action leads to its own coins to spawn for several time steps, and standard agent collects mostly the coins of the spawn-biased during this time period. IQL with IA shows a better result to IQL yet Fair\&LocalIA converges to fully cooperative behaviors much faster than its original. Similarly, IQL with Fair\&LocalSVO leads to more cooperative policies compared to IQL with SVO in this setting.

\figref \ref{fig:asymmetric_commons_harvest__as_5adv_VS_5dis_vs_fair_local} shows that, in \texttt{Harvest} with \textit{asymmetry in zapping}, IQL again converges to a defective setting with low peace where wide-zap agents are zapping more effectively. While showing higher returns, higher peace and higher sustainability compared to IQL, IQL with IA amplifies the gap between the wide-zap and standard agents in terms of these metrics. On the other hand, IQL w/ Fair\&Local removes the gap with much less defective policies leading to higher peace. A similar result is seen for SVO and Fair\&LocalSVO, where our proposed version makes the wide-zap and standard agents perform more similarly in terms of cooperation.

Additional results analyzing the convergence of the proposed method’s key features are presented in Appendix \ref{sec:additional_results}.


\section{Discussion \& Future Work}

In this paper, we introduced asymmetry into sequential social dilemmas, by creating differences between agents in terms of reward functions and action effects, a crucial setting for understanding how cooperation can emerge when individual and group interests conflict. Our results showed that asymmetry naturally leads to distinct behaviors across different agent types. While fairness-based methods such as IA and SVO modify agents' reward functions to encourage prosocial behavior, in our experiments, they failed to fully address the inherent asymmetry, sometimes creating incentives to defect. In addition, these methods require constant access to other agents' returns during training.

We proposed modifications to IA and SVO, including normalization for agent potential, agent-based weighting to account for asymmetry, and local social feedback. The resulting Fair\&LocalIA and Fair\&LocalSVO methods handle asymmetry more effectively, promoting higher levels of cooperation across all agent types and avoiding defection, while relying only on local information.

Future work could explore other forms of asymmetry, such as differential access to resources or varied learning capabilities. Additionally, agent-based weights could be learned automatically during training, eliminating the need to specify them beforehand.



\begin{acks}
This research is financially supported by the Scientific and Technological Research Council of Turkey through BIDEB 2219 International Postdoctoral Research Scholarship Program.
\end{acks}



\bibliographystyle{ACM-Reference-Format} 
\bibliography{references}


\appendix

\section{Verification of the Asymmetric Environments}
\label{sec:env_verification}

\begin{figure*}
    \centering
    \begin{subfigure}[b]{0.23\textwidth}
        \centering
        \includegraphics[width=1\textwidth]{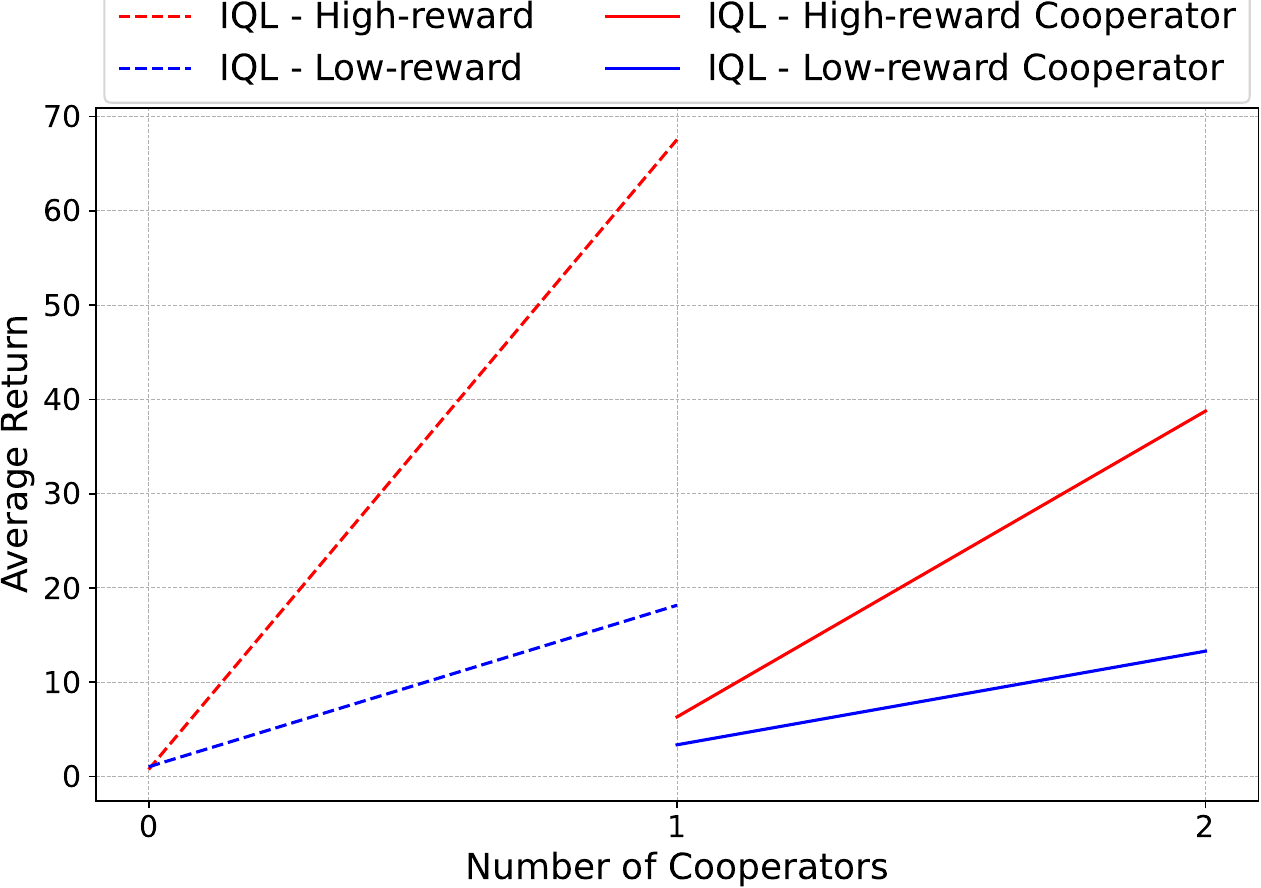}
        \caption{\texttt{Coins} with \textit{asymmetry in coin rewards}}
        \label{fig:asymmetric_coins__rf_1adv_1dis_schelling_diagram}
    \end{subfigure} 
    \quad
    \begin{subfigure}[b]{0.23\textwidth}
        \centering
        \includegraphics[width=1\textwidth]{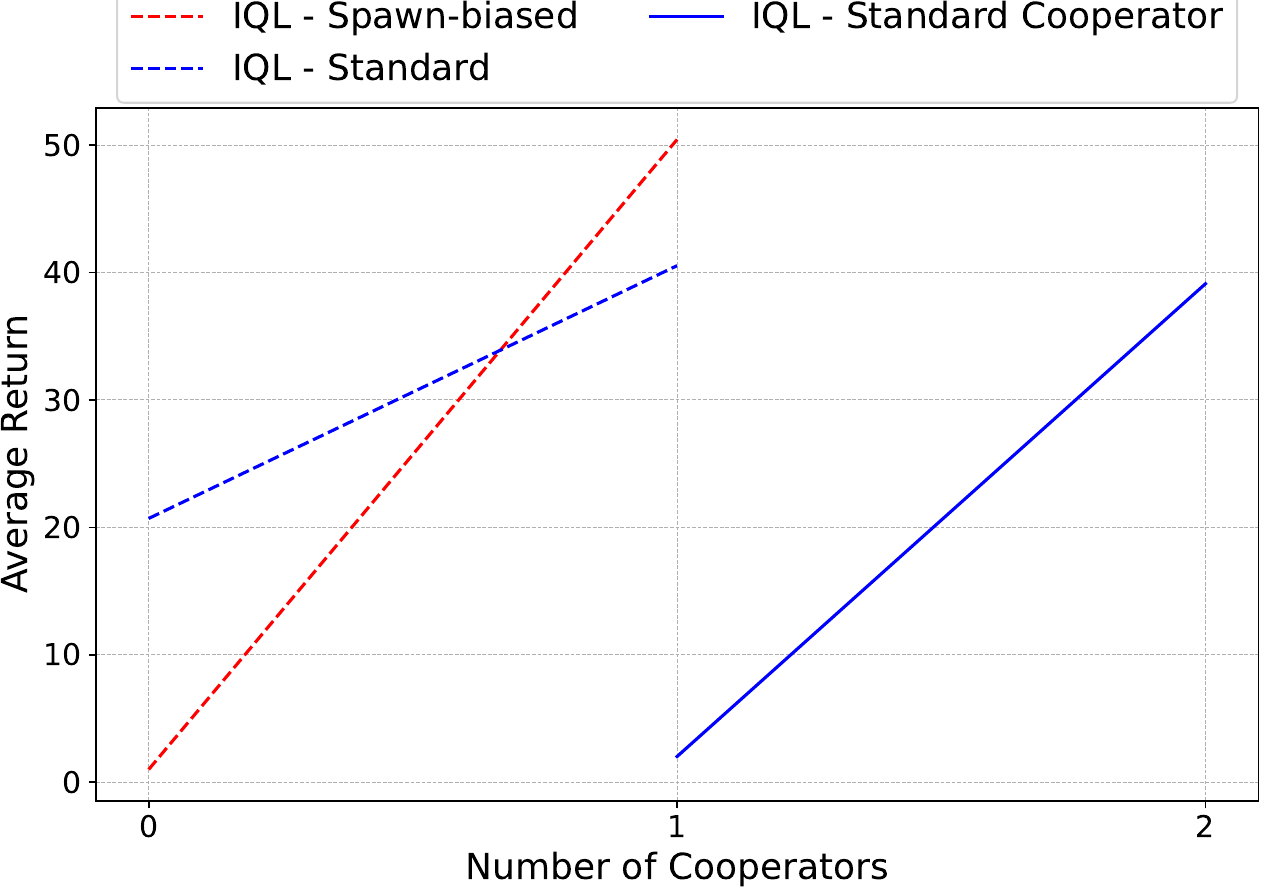}
        \caption{\texttt{Coins} with \textit{asymmetry in coin spawn}}
        \label{fig:asymmetric_coins__as_1adv_1dis_schelling_diagram}
    \end{subfigure} 
    \quad
    \begin{subfigure}[b]{0.23\textwidth}
        \centering
        \includegraphics[width=1\textwidth]{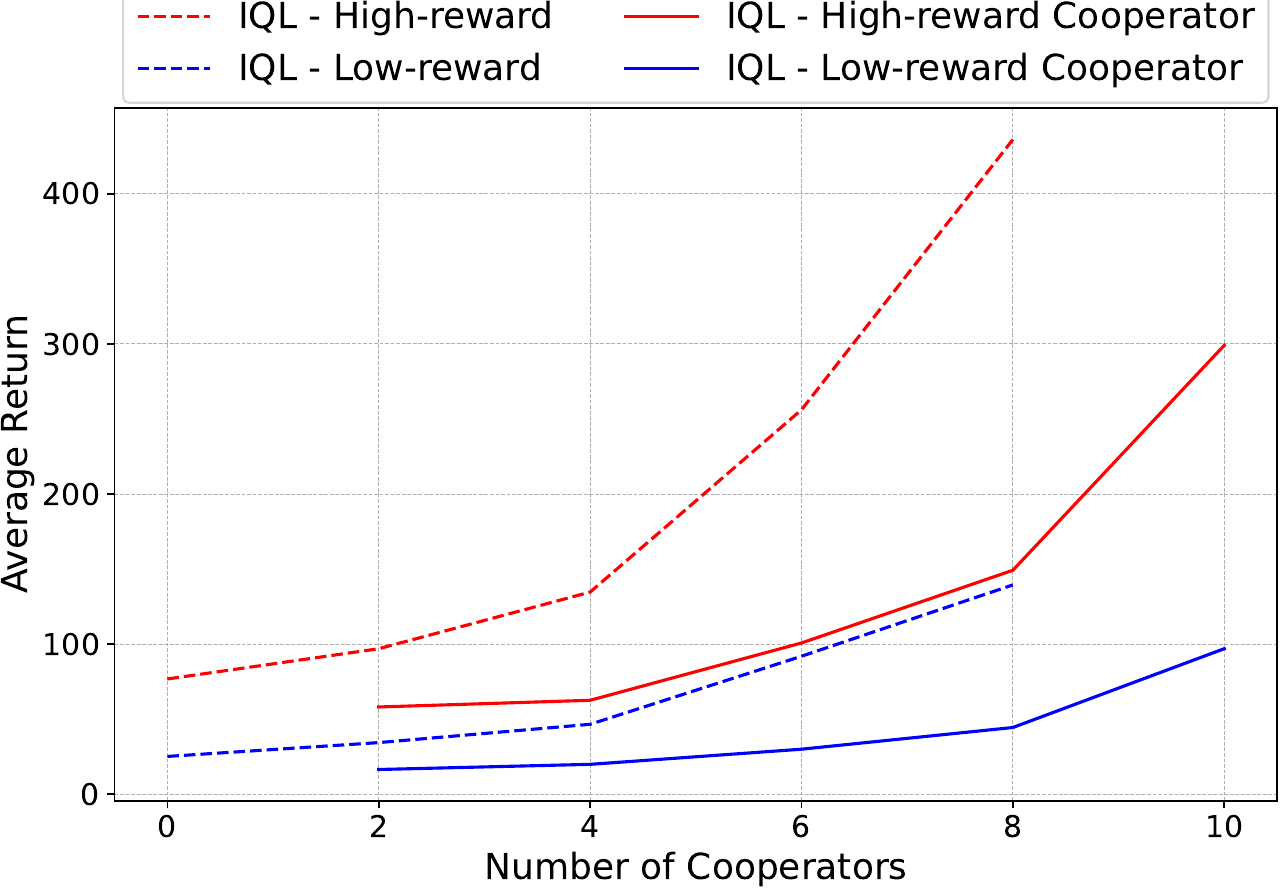}
        \caption{\texttt{Harvest} with \textit{asymmetry in apple rewards}}
        \label{fig:asymmetric_commons_harvest__rf_5adv_5dis_schelling_diagram}
    \end{subfigure}
    \quad
    \begin{subfigure}[b]{0.23\textwidth}
        \centering
        \includegraphics[width=1\textwidth]{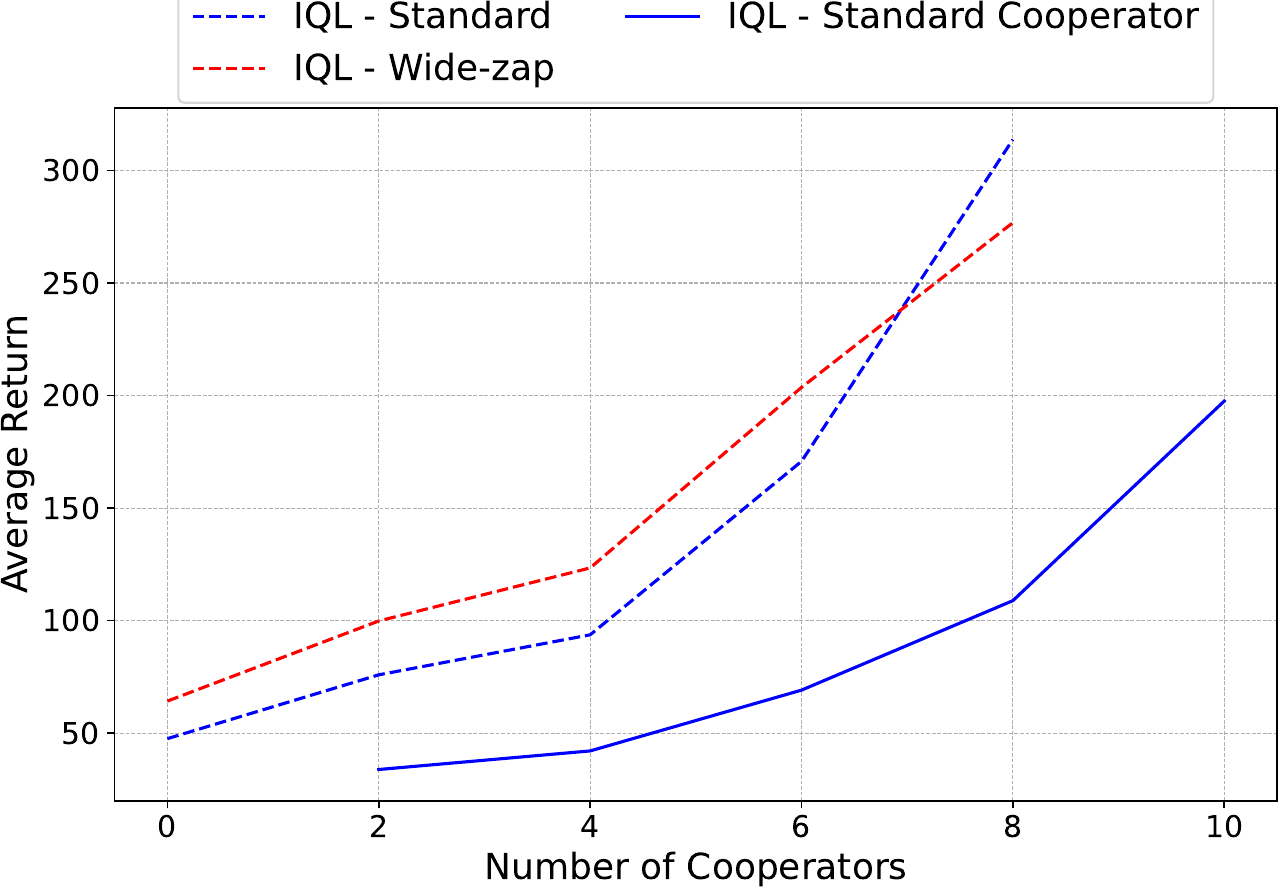}
        \caption{\texttt{Harvest} with \textit{asymmetry in zap radius}}
        \label{fig:asymmetric_commons_harvest__as_5adv_5dis_schelling_diagram}
    \end{subfigure}
    \caption{Schelling diagrams for the asymmetric versions of the environments. Note that there is no asymmetry when defection is prevented in \texttt{Coins} with \textit{asymmetry in coin spawn} and \texttt{Harvest} with \textit{asymmetry in zap radius} and thus the cooperating agents are represented as a single line. The results are averaged over 5 experiments where each experiment evaluates the policies for 10 episodes.}
    \label{fig:asymmetric_schelling_diagrams}
    \Description{}
\end{figure*} 

To verify that the asymmetric versions still exhibit the characteristics of a social dilemma (conditions C1 to C4), we ran experiments with varying numbers of cooperating agents and plotted Schelling diagrams \cite{schelling1973hockey} (\figref \ref{fig:asymmetric_schelling_diagrams}) to show that defection leads to higher returns compared to cooperation. Since there are multiple agent types, the diagrams show cooperation and defection returns separately for each type. In \texttt{Coins}, cooperators are restricted from collecting a mismatch coin. In \texttt{Harvest}, cooperation is enforced by preventing the zap action and the collection of apples from low-density areas. 

In \texttt{Coins}, \figref \ref{fig:asymmetric_coins__rf_1adv_1dis_schelling_diagram} and \ref{fig:asymmetric_coins__as_1adv_1dis_schelling_diagram} show that, as the proportion of cooperative policies increases, defective policies yield higher returns for both agent types compared to their cooperative counterparts. In the variant with asymmetry in coin spawn, the relative advantage shifts between agent types as cooperation levels rise. Both versions of the environment exhibit the defining characteristics of a social dilemma.

In \texttt{Harvest}, \figref \ref{fig:asymmetric_commons_harvest__rf_5adv_5dis_schelling_diagram} demonstrates that, for both high-reward and low-reward agents, cooperative behavior results in lower returns than defection, confirming that this environment retains the key features of a social dilemma. Similarly, \figref \ref{fig:asymmetric_commons_harvest__as_5adv_5dis_schelling_diagram} shows that cooperation again leads to lower returns, reinforcing the presence of a social dilemma in this variant as well.

\section{Implementation Details}
\label{sec:implementation}

In both environments, a single agent can only observe a small section ($11 \times 11$ view, 8 pixels for each tile, resulting in $88 \times 88 \times 3$ RGB images as observations) around itself and is equipped with the actions of \textit{staying put}, \textit{going forward}, \textit{turning left}, and \textit{turning right}. Another agent becomes visible when it is in this view of $11 \times 11$. In \texttt{Harvest} environment, there is an additional action of \textit{zapping} that removes the zapped agent from the environment for 25 time steps. In \texttt{Coins} environment, there is at most 1 coin in the environment at a time, an uncollected coin is removed after 50 time steps, the regrowth rate is 0.1 and each agent's view cover the entire grid.

All environments in this study, including the modified versions, build on the Melting Pot framework \cite{leibo2021meltingpot, agapiou2022melting}, which is optimized for efficient training.

\begin{figure*}
    \centering
    \begin{subfigure}[b]{0.12\textwidth}
        \centering
        \includegraphics[width=\textwidth]{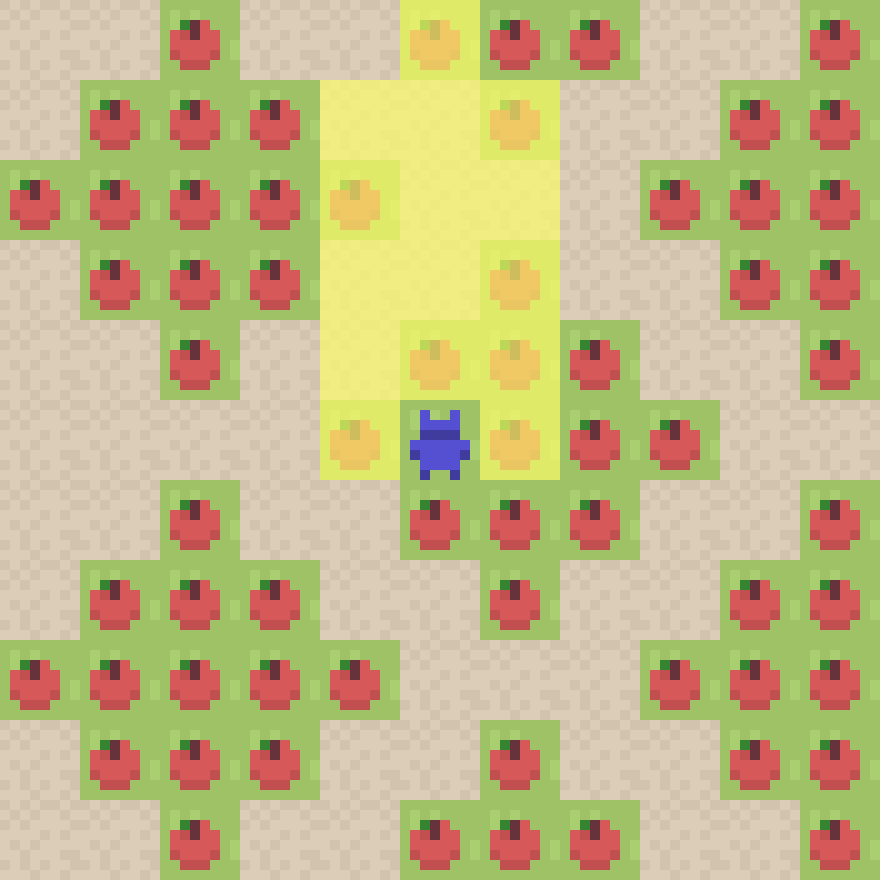}
        \caption{Standard}
    \end{subfigure} 
    \quad
    \begin{subfigure}[b]{0.12\textwidth}
        \centering
        \includegraphics[width=\textwidth]{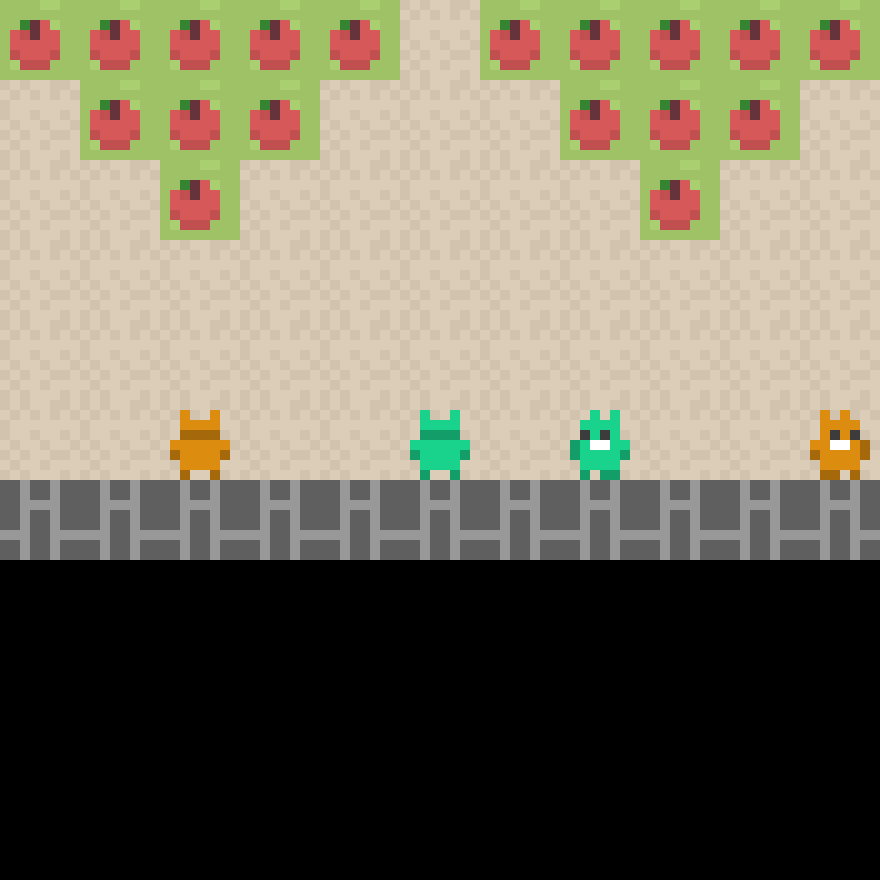}
        \caption{High-reward}
    \end{subfigure}
    \quad
    \begin{subfigure}[b]{0.12\textwidth}
        \centering
        \includegraphics[width=\textwidth]{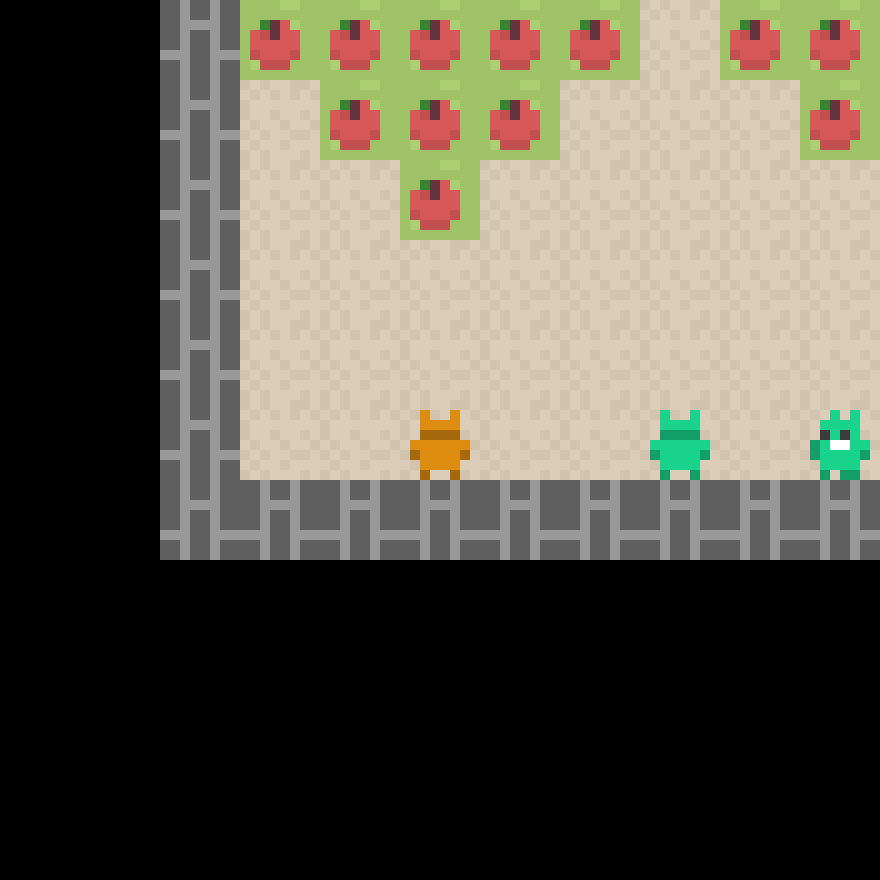}
        \caption{Low-reward}
    \end{subfigure}  
    \quad
    \begin{subfigure}[b]{0.12\textwidth}
        \centering
        \includegraphics[width=\textwidth]{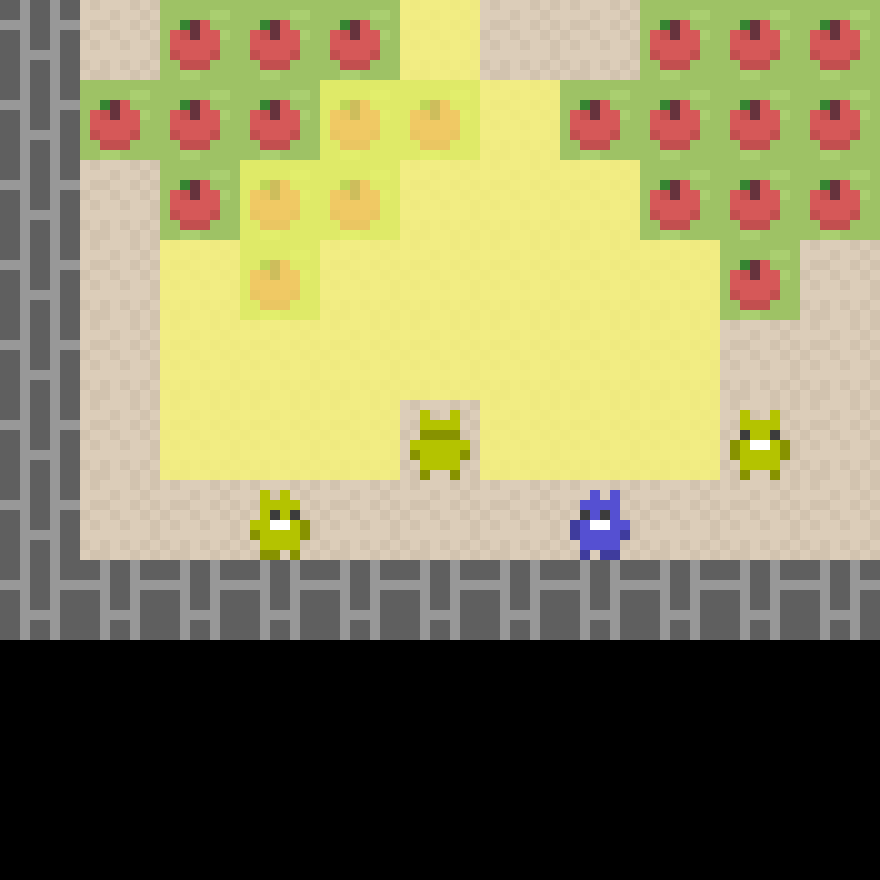}
        \caption{Wide-zap}
    \end{subfigure}    
    \caption{(a) \texttt{Harvest} environment with 10 standard agents and types of agents (b)-(e) used within this environment.}
    \label{fig:harvest_environment}
    \Description{}
\end{figure*}   

The agents are trained using the Independent DQN (IQL) implementation from the BenchMARL framework \cite{bettini2024benchmarl}, which is fully compatible with the Melting Pot environments.

The following learning parameters is selected for the experiments. Each agent model includes two convolutional layers with 32 and 128 cells, kernel sizes of 8 and 11, strides of 8 and 1 without padding. These layers are followed by a feed forward layer of 64 cells and another LSTM layer of 64 cells and a final feed forward layer of 64 cells. All layers used Rectified Linear Units (ReLU) as the activation function. The agents used separate replay buffers of 20000 steps, discount rate as 0.99, learning rate as $5 \times 10^{-5}$ and epsilon-greedy action selection with epsilon decaying from 0.8 to 0.1 within the first million steps. The test period was set to 50000 steps with 10 episodes in each evaluation iteration for both environments. Experimental results are aggregated over 10 trials where each episode lasts for 500 and 1000 steps in \texttt{Coins} and \texttt{Harvest}, respectively. The shaded areas represent 95\% boostrapped confidence intervals.

\begin{figure*}
    \centering
    \begin{subfigure}[b]{0.35\textwidth}
        \centering
        \includegraphics[width=0.25\textheight]{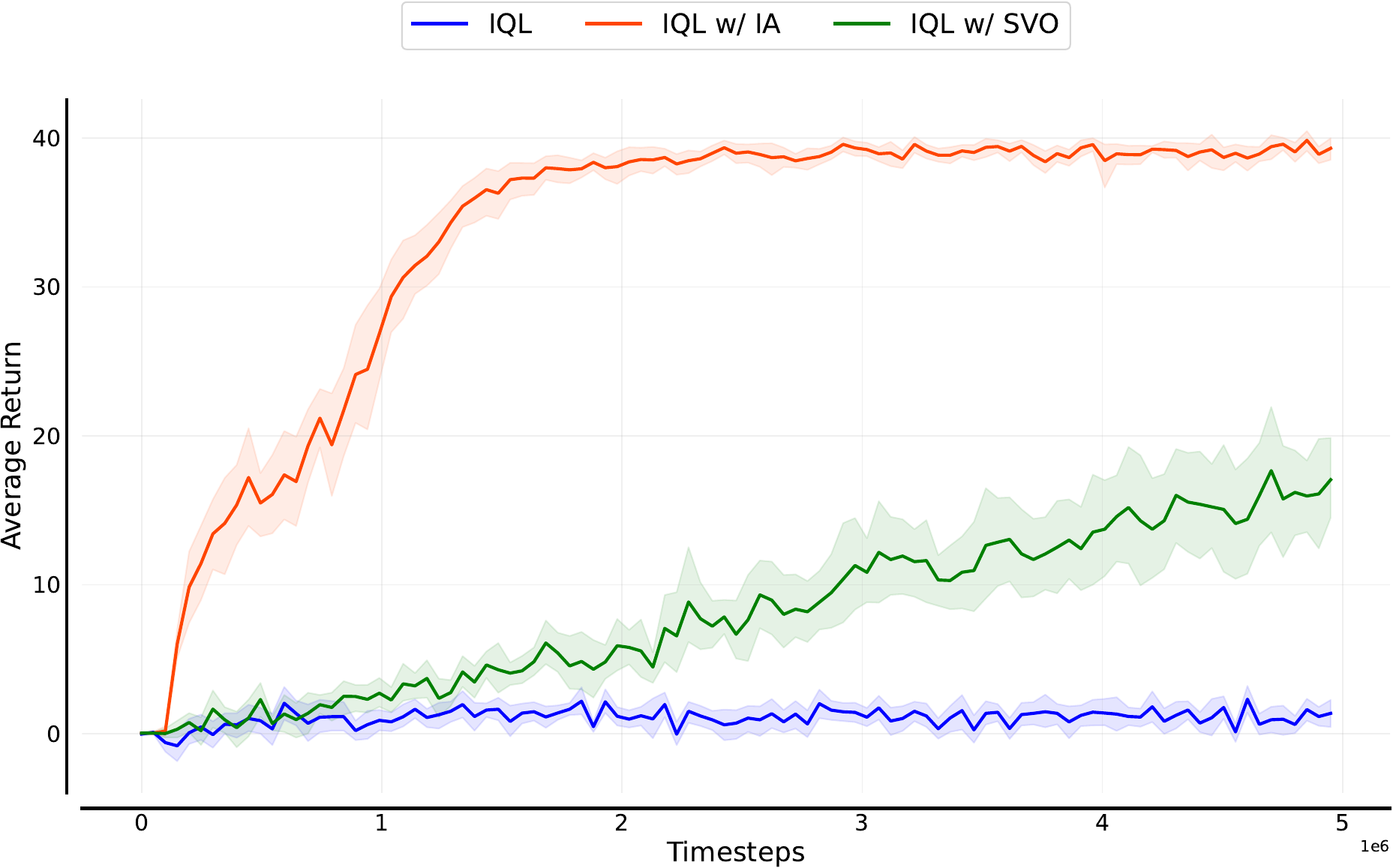}
    \end{subfigure}    
    \begin{subfigure}[b]{0.35\textwidth}
        \centering
        \includegraphics[width=0.25\textheight]{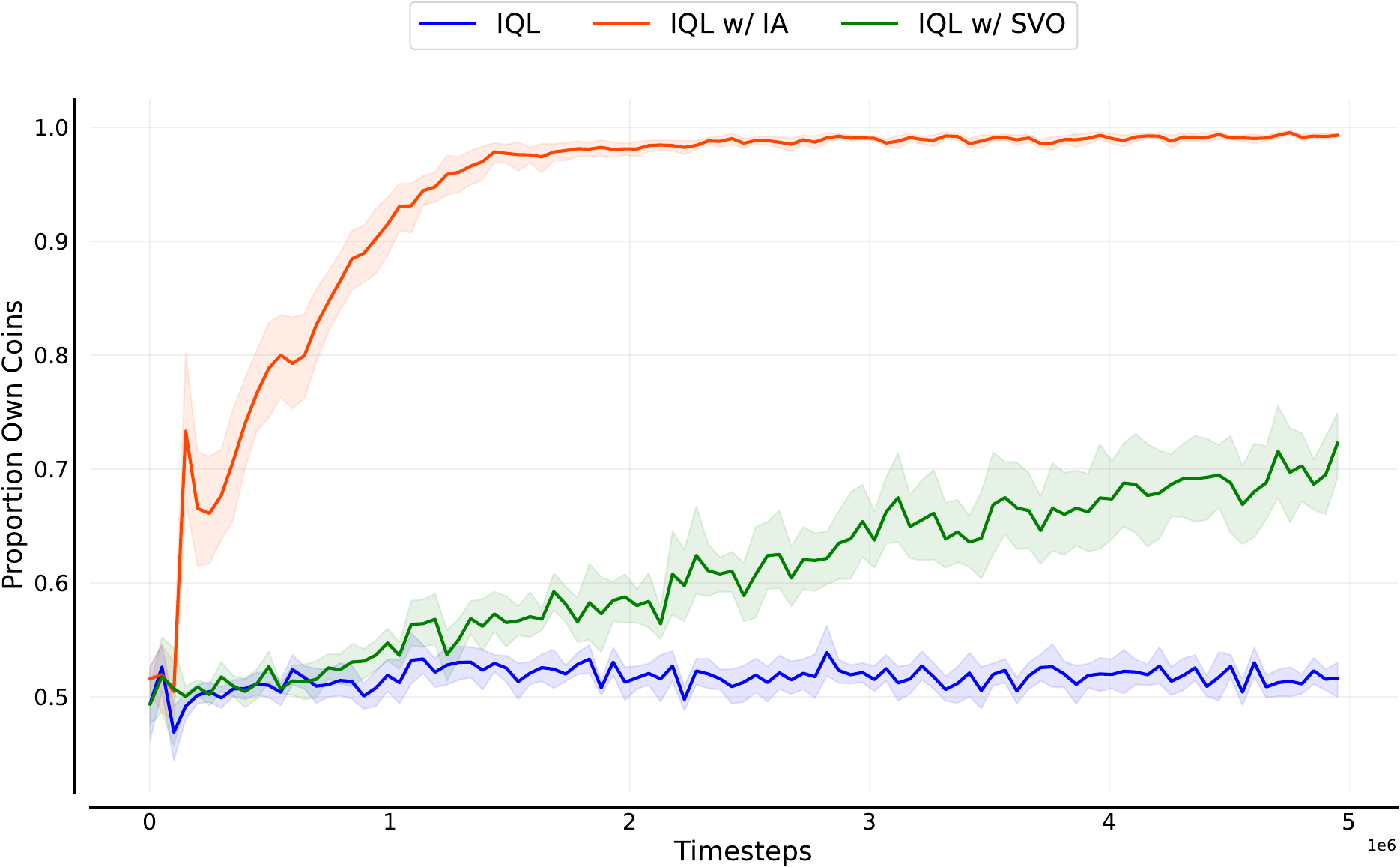}
    \end{subfigure}   
    \caption{Performance of IA and SVO in symmetric \texttt{Coins} environment. Even the best hyperparameters of SVO cannot perform as well as IA in this environment. }
    \label{fig:iql_asymmetric_coins__default_VS_SVO_IA}
    \Description{}
\end{figure*}

\begin{figure*}
    \centering
    \begin{subfigure}[b]{0.35\textwidth}
        \centering
        \includegraphics[width=0.25\textheight]{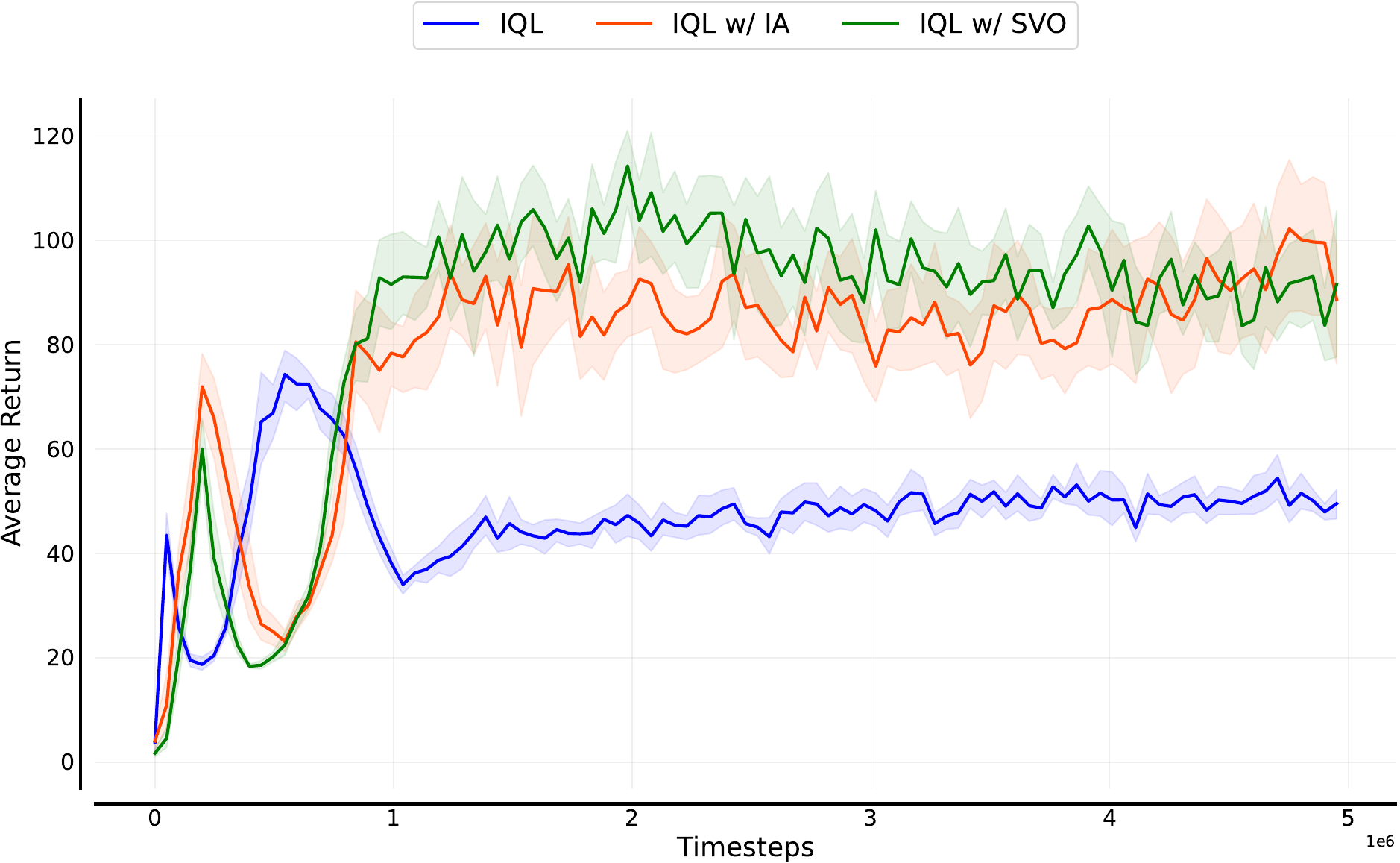}
    \end{subfigure}    
    \begin{subfigure}[b]{0.35\textwidth}
        \centering
        \includegraphics[width=0.25\textheight]{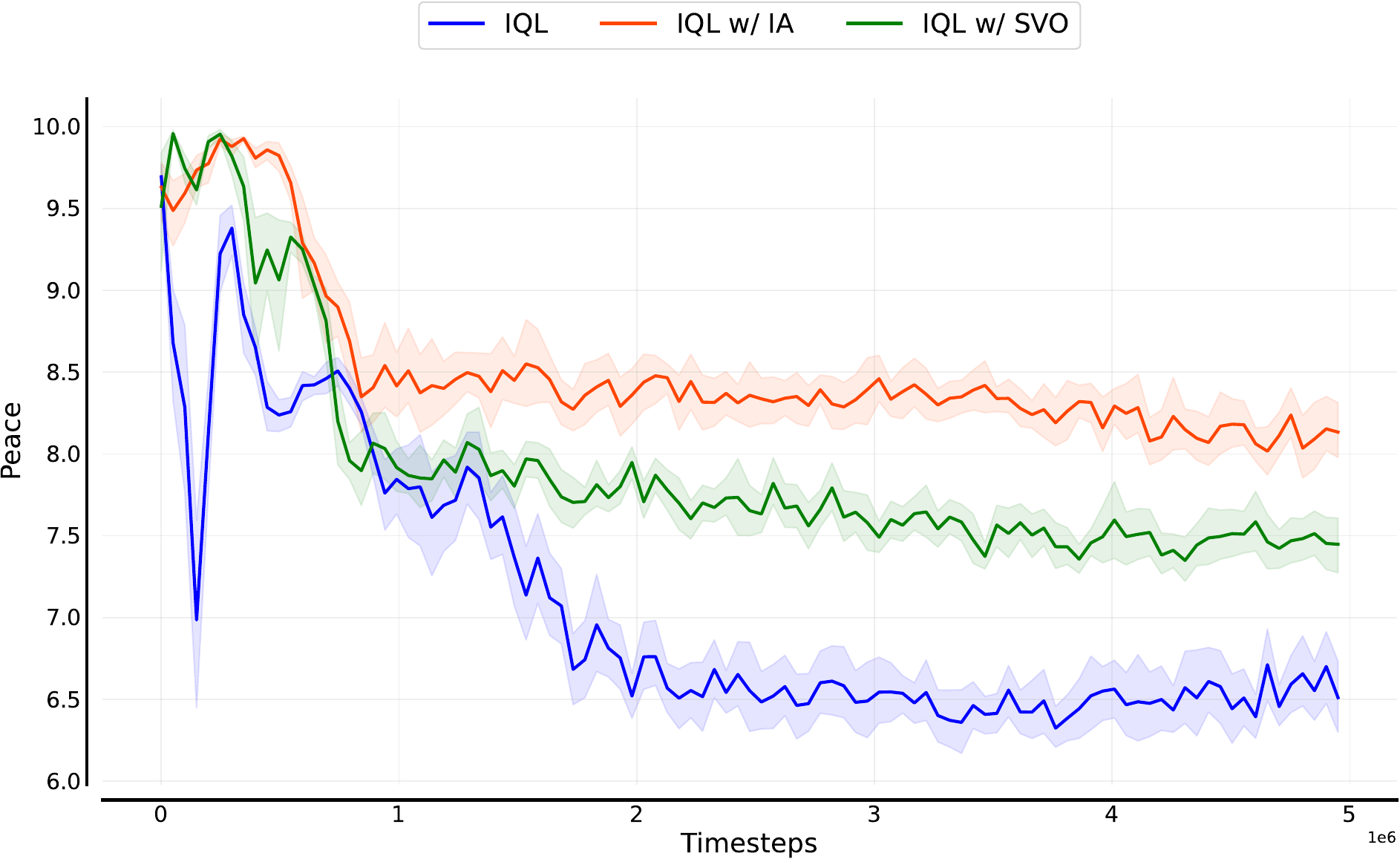}
    \end{subfigure}  
    \begin{subfigure}[b]{0.35\textwidth}
        \centering
        \includegraphics[width=0.25\textheight]{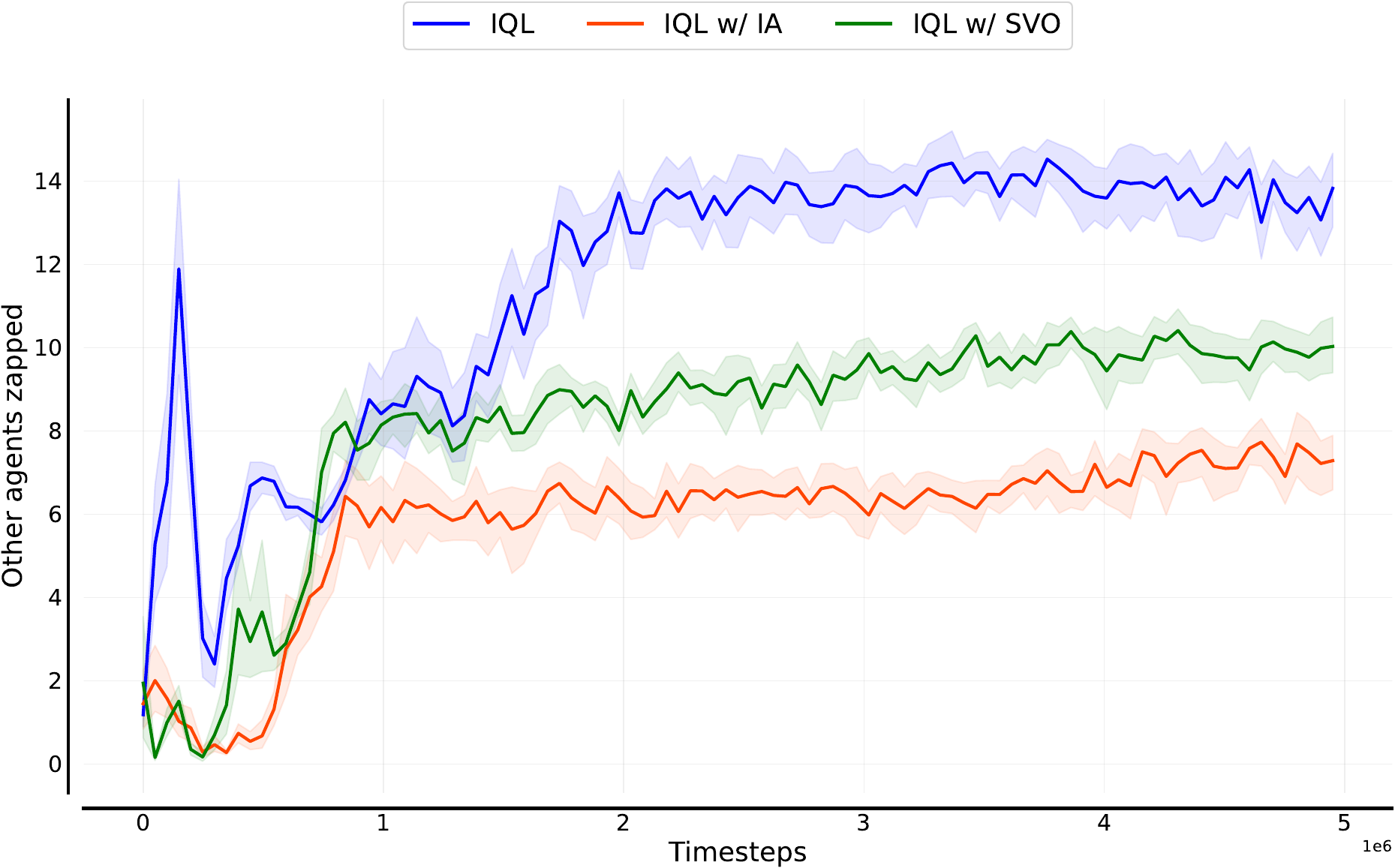}
    \end{subfigure}  
    \begin{subfigure}[b]{0.35\textwidth}
        \centering
        \includegraphics[width=0.25\textheight]{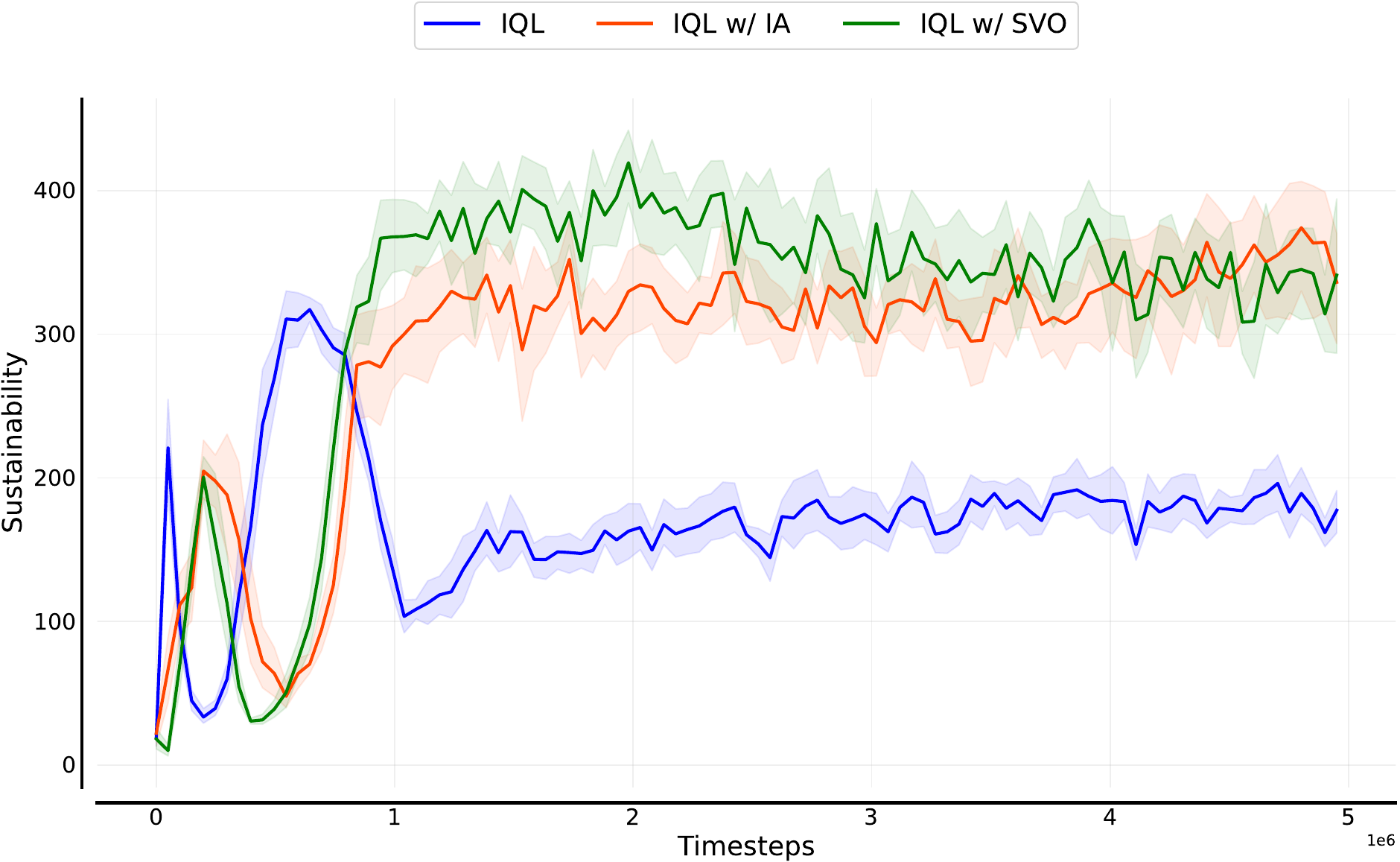}
    \end{subfigure}     
    \caption{Performance of  IA and SVO in symmetric \texttt{Harvest} environment. Both show outperforming performance compared to baseline IQL.}
    \label{fig:iql_asymmetric_commons_harvest__default_VS_SVO_IA}
    \Description{}
\end{figure*}

Starting from their suggested values in their respective papers, a grid search is employed to find the best hyperparameters for the baseline methods of IA and SVO. The performance of IA and SVO are verified in the symmetric versions of the environments with the found hyperparameters and shown in \figref \ref{fig:iql_asymmetric_coins__default_VS_SVO_IA} and \ref{fig:iql_asymmetric_commons_harvest__default_VS_SVO_IA}. The decay of the temporally smoothed rewards, $\lambda$ is set to $0.9$ in all cases. The rest of the hyperparameters are given in Tables \ref{tab:ia_hyperparameters} and \ref{tab:svo_hyperparameters}.

\begin{table*}
    \centering
    \begin{tabular}{c|c|c|c|c|c|c|c}
         & $\alpha$ & $\beta$ & $\phi_{low-reward}$ & $\phi_{standard}$ & $\phi_{high-reward}$ & $\phi_{wide-zap}$ & $\phi_{spawn-biased}$\\ \hline 
        \texttt{Coins} & 0.05 & 0.1 & 4.0 & 13.5 & 12.0 & - & 8.0 \\
        \texttt{Harvest} & 3 & 0.05 & 4.0 & 4.0 & 6.0 & 6.0 & - \\
    \end{tabular}
    \caption{IA and Fair\&LocalIA hyperparameters}
    \label{tab:ia_hyperparameters}
\end{table*}

\begin{table*}
    \centering
    \begin{tabular}{c|c|c|c|c|c|c|c}
         & $\theta^{SVO}$ & $w$ & $\phi_{low-reward}$ & $\phi_{standard}$ & $\phi_{high-reward}$ & $\phi_{wide-zap}$ & $\phi_{spawn-biased}$ \\ \hline 
        \texttt{Coins} & $45\degree$ & 0.004 & 2.0 & 4.5 & 6.0 & - & 1.5 \\
        \texttt{Harvest} & $45\degree$ & 0.02 & 1.0 & 1.0 & 1.5 & 1.5 & - \\
    \end{tabular}
    \caption{SVO and Fair\&LocalSVO hyperparameters}
    \label{tab:svo_hyperparameters}
\end{table*}

\section{Convergence of Local Estimates and Smoothed Rewards}
\label{sec:additional_results}

Additionally, we track method specific metrics such as:  \textit{the average age of agent estimates} for the temporally smoothed rewards as 
\begin{align*}
    \hat{\tau} = \frac{1}{N}\frac{1}{T}\sum_{t = 0}^T \sum_{i \in \mathcal{N}} \sum_{j \neq i} (t - \tau_{i,j}^t),
\end{align*}
\textit{the average range of temporally smoothed rewards} as
\begin{align*}
    e^{avg} = \frac{1}{N} \sum_{i \in \mathcal{N}} (e_i^{max} - e_i^{min}).
\end{align*}

\begin{figure*}[h]
    \centering
    \begin{subfigure}[b]{0.44\textwidth}
        \centering
        \includegraphics[width=0.22\textheight]{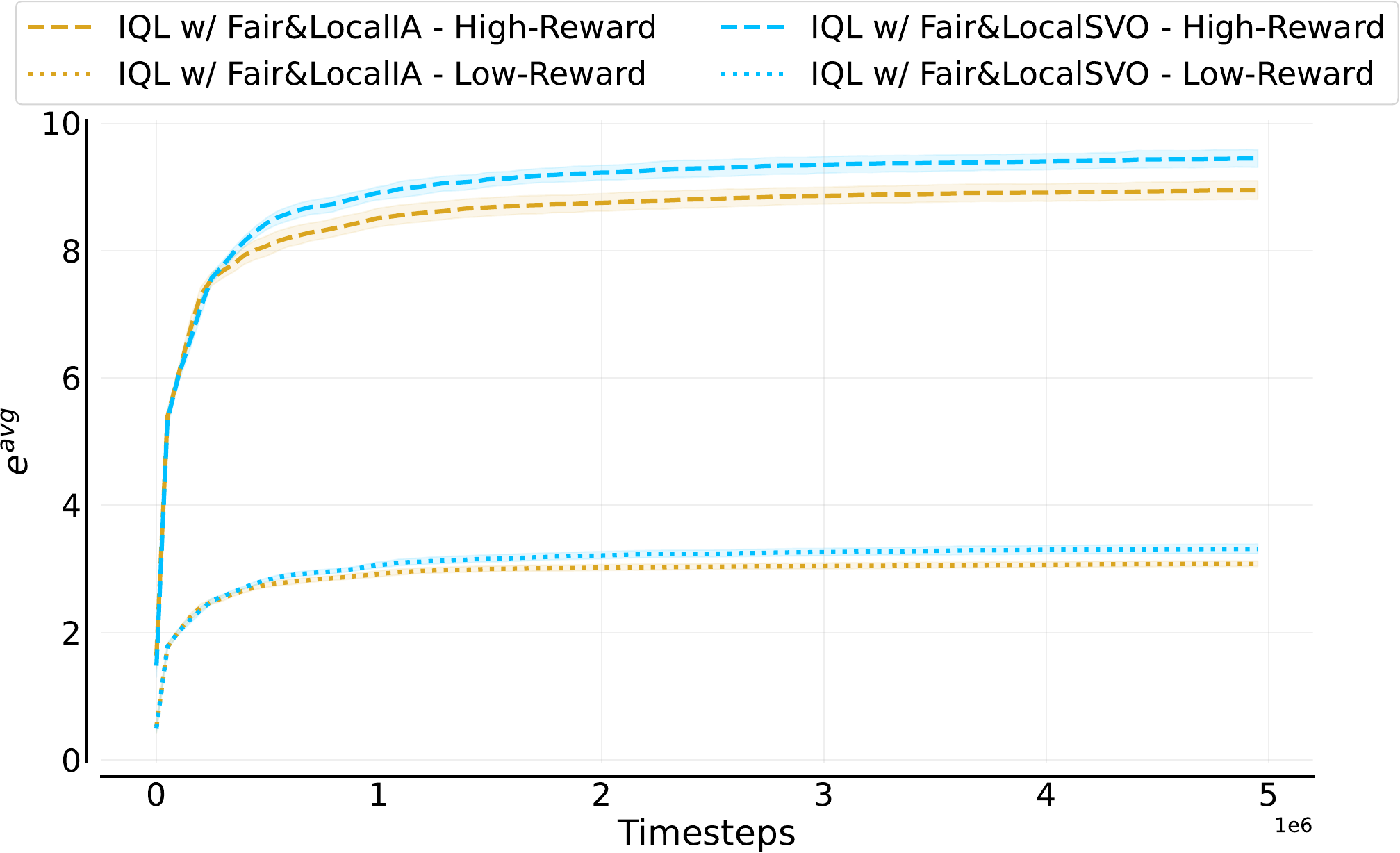}
    \end{subfigure} 
    \begin{subfigure}[b]{0.44\textwidth}
        \centering
        \includegraphics[width=0.22\textheight]{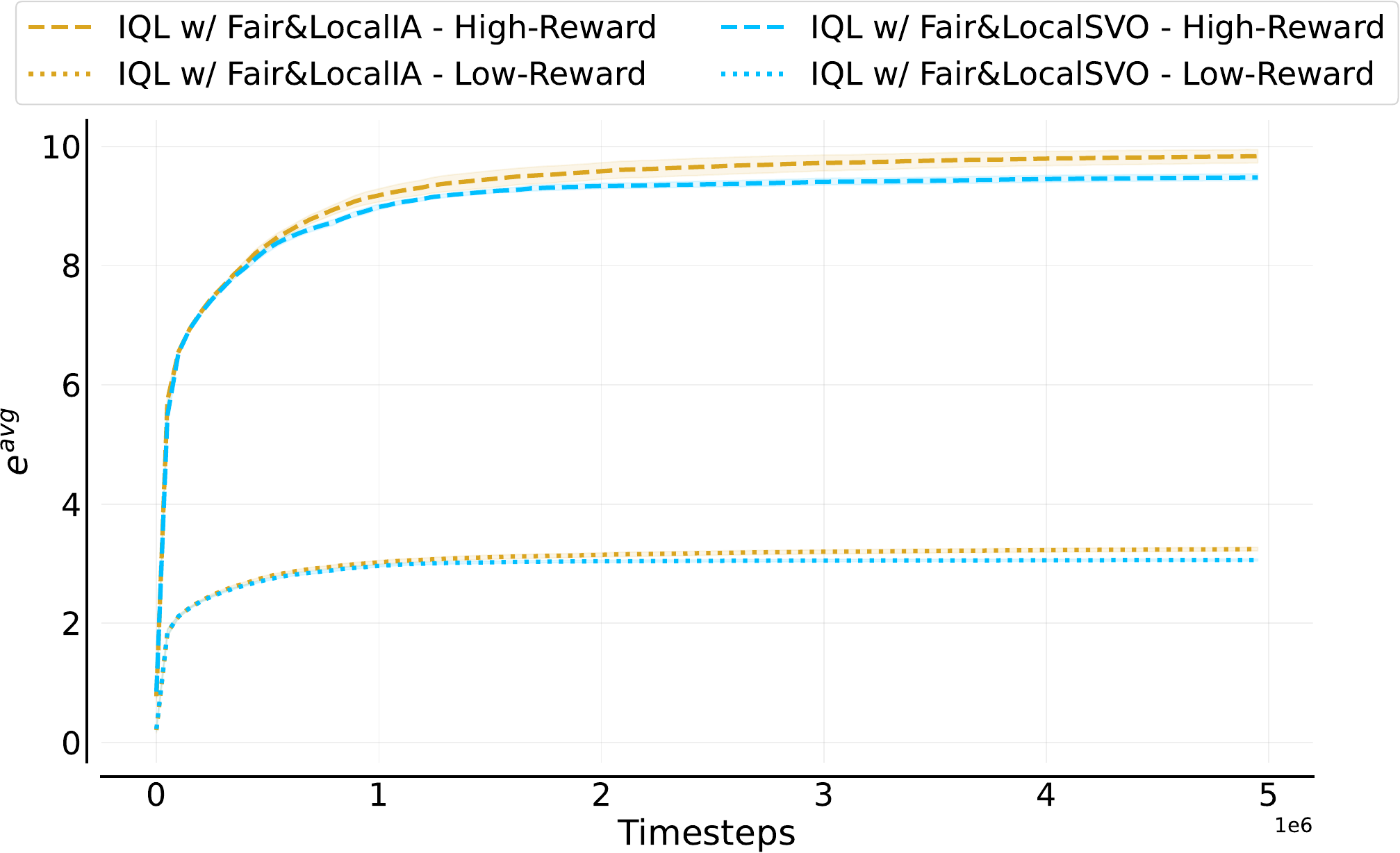}
    \end{subfigure}    
    \caption{\textit{The average range of temporally smoothed rewards} for Fair\&Local versions of IA and SVO in \texttt{Coins} environment with \textit{asymmetry in coin rewards} and \texttt{Harvest} with \textit{asymmetry in apple rewards}. The value converge quickly during training.}
    \label{fig:tsr_range_coins__rf_1adv_1dis_commons_harvest__rf_5adv_5dis}
    \Description{}
\end{figure*}

\begin{figure*}
    \centering
    \begin{subfigure}[b]{0.44\textwidth}
        \centering
        \includegraphics[width=0.22\textheight]{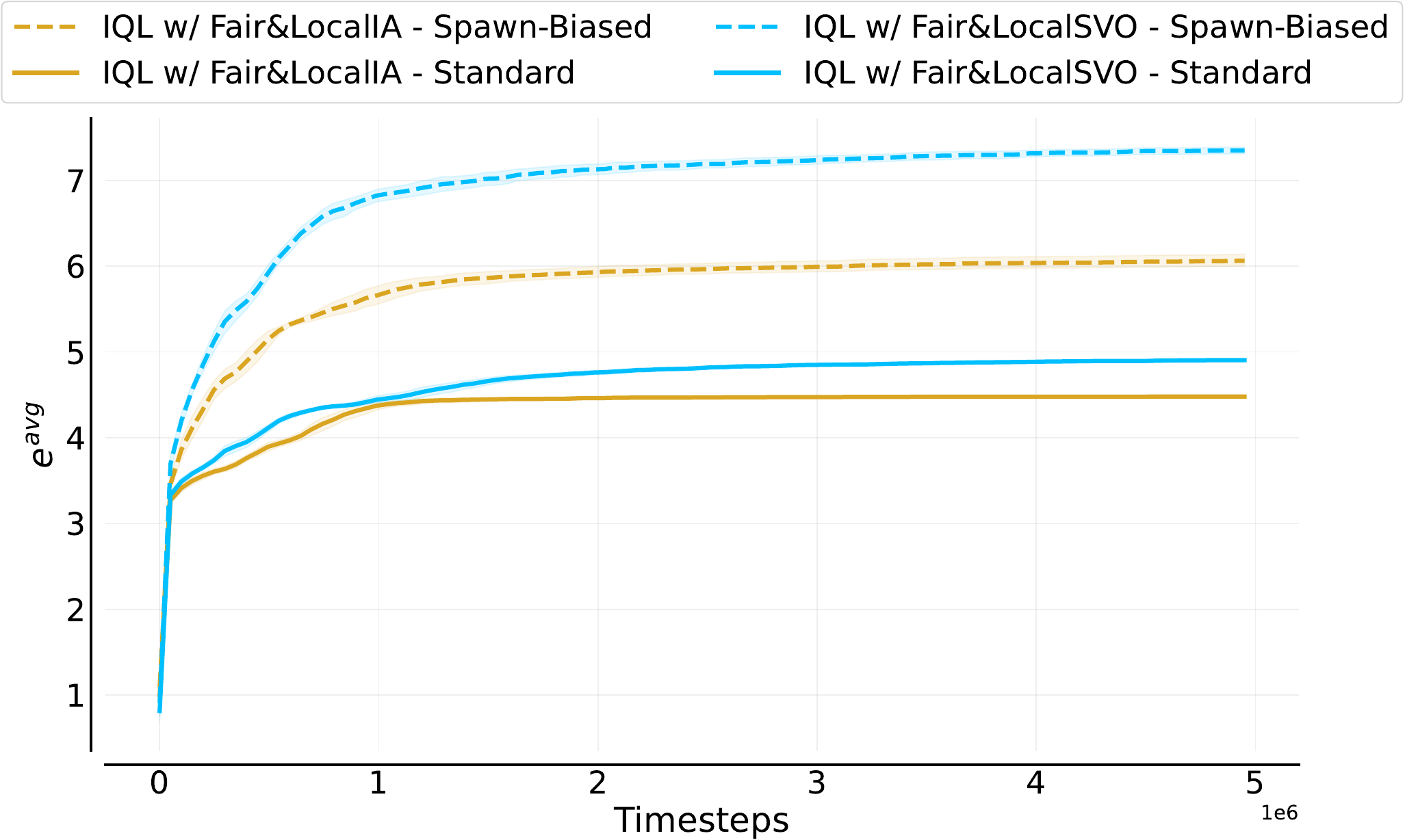}
    \end{subfigure} 
    \begin{subfigure}[b]{0.44\textwidth}
        \centering
        \includegraphics[width=0.22\textheight]{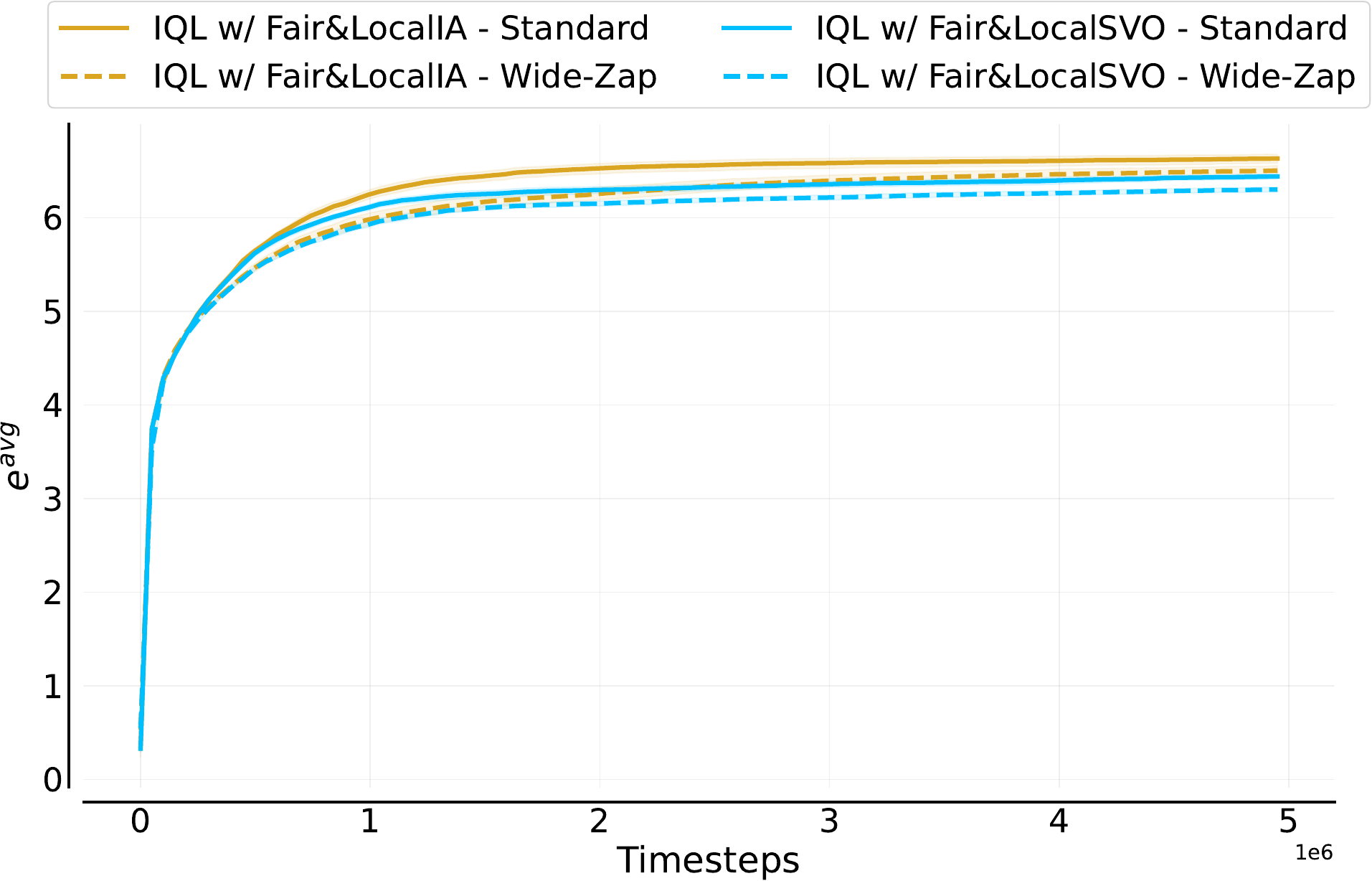}
    \end{subfigure}    
    \caption{\textit{The average range of temporally smoothed rewards} for Fair\&Local versions of IA and SVO in \texttt{Coins} environment with \textit{asymmetry in coin spawn} and \texttt{Harvest} environment with \textit{asymmetry in zap radius}.}
    \label{fig:tsr_range_coins_as__1adv_1dis_commons_harvest__as_5adv_5dis}
    \Description{}
\end{figure*} 

\begin{figure*}
    \centering
    \begin{subfigure}[b]{0.44\textwidth}
        \centering
        \includegraphics[width=0.22\textheight]{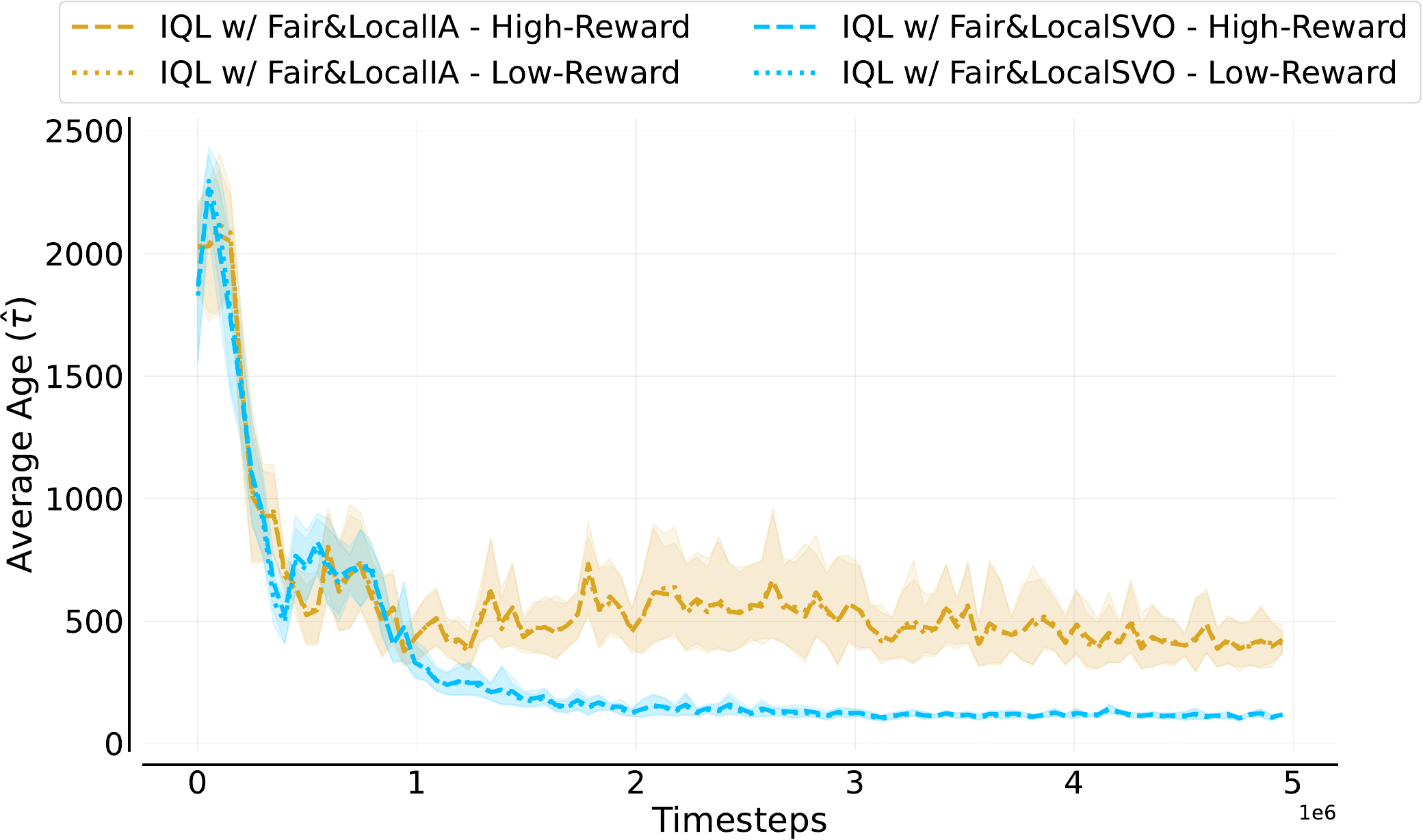}
    \end{subfigure} 
    \begin{subfigure}[b]{0.44\textwidth}
        \centering
        \includegraphics[width=0.22\textheight]{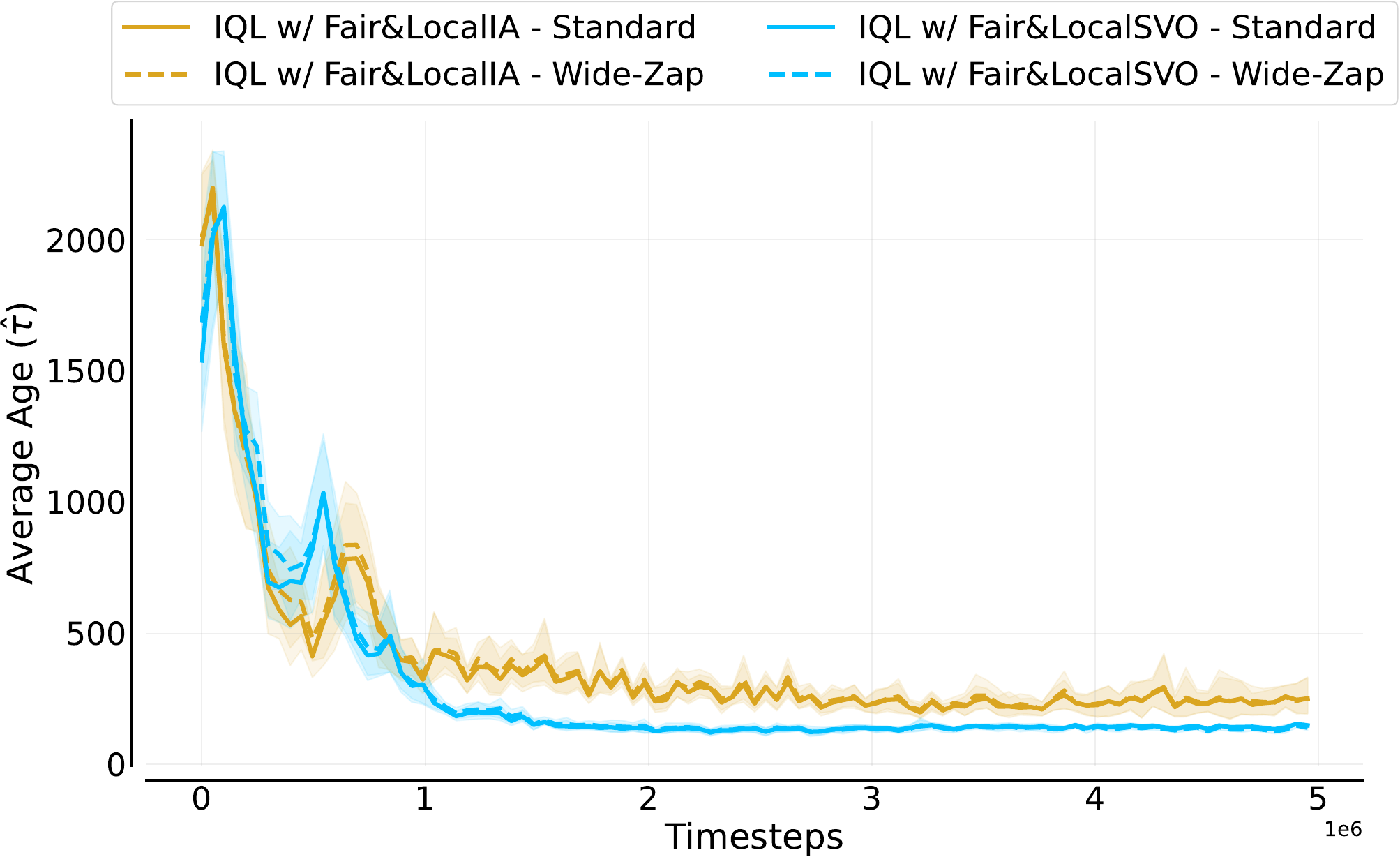}
    \end{subfigure}    
    \caption{\textit{The average age of agent estimates} for the temporally smoothed rewards for Fair\&Local versions of IA and SVO in \texttt{Harvest} environment with \textit{asymmetry in apple rewards} and \texttt{Harvest} environment with \textit{asymmetry in zap radius}.}
    \label{fig:asymmetric_commons_harvest__average_age}
    \Description{}
\end{figure*} 

In the case of asymmetry in rewards, \figref \ref{fig:tsr_range_coins__rf_1adv_1dis_commons_harvest__rf_5adv_5dis} also shows that the range of temporally smoothed rewards, used in Eq. \ref{eq:tsr_norm}, converge quickly during training and show the difference in rewards clearly.

It is shown in \figref \ref{fig:tsr_range_coins_as__1adv_1dis_commons_harvest__as_5adv_5dis} that spawn-biased agent in \texttt{Coins} with \textit{asymmetry in coin spawn} show a larger temporally smoothed reward range as their defection causes defection of standard agent whereas there is no clear difference in terms of these ranges in \texttt{Harvest} with \textit{asymmetry in zap radius}. Still, these values converge quickly during training, making Eq. \ref{eq:tsr_norm} applicable in these settings.

In \texttt{Harvest} environment, an agent's view does not cover the entire grid, so the localization of the social feedback becomes a key factor in the performances of Fair\&Local versions of IA and SVO. \figref \ref{fig:asymmetric_commons_harvest__average_age} shows that average age of agent estimates for temporally smoothed rewards converge to low values quickly, where Fair\&LocalSVO shows much lower values. This is due to the dynamic nature of the environment and the communication between agents according to Algorithm \ref{alg:tsr_update}.

\end{document}